\documentclass{article}

\usepackage{microtype}
\usepackage{graphicx}
\usepackage{subfigure}
\usepackage{booktabs} 

\usepackage{hyperref}

\usepackage[accepted]{icml2024}

\usepackage{amsmath}
\usepackage{amssymb}
\usepackage{mathtools}
\usepackage{amsthm}

\usepackage[capitalize,noabbrev]{cleveref}

\theoremstyle{plain}

\theoremstyle{definition}

\theoremstyle{remark}

\usepackage{array,multirow,booktabs}

\usepackage[textsize=tiny]{todonotes}

\icmltitlerunning{Examining Modality Incongruity in Multimodal Federated Learning}

\begin{document}

\twocolumn[
\icmltitle{Examining Modality Incongruity in Multimodal Federated Learning \\for Medical Vision and Language-based Disease Detection}




\begin{icmlauthorlist}
\icmlauthor{Pramit Saha}{yyy}
\icmlauthor{Divyanshu Mishra}{yyy}
\icmlauthor{Felix Wagner}{yyy}
\icmlauthor{Konstantinos Kamnitsas}{yyy}
\icmlauthor{J. Alison Noble}{yyy}

\end{icmlauthorlist}

\icmlaffiliation{yyy}{Department of Engineering Science, University of Oxford, United Kingdom}

\icmlcorrespondingauthor{Pramit Saha}{pramit.saha@eng.ox.ac.uk}

\icmlkeywords{Federated Learning, Multimodal Learning, Missing modality, Modality Incongruity, Chest X-Ray, Radiology report}
\vskip 0.3in]



\printAffiliationsAndNotice{}  

\begin{abstract}
Multimodal Federated Learning (MMFL) utilizes multiple modalities in each client to build a more powerful Federated Learning (FL) model than its unimodal counterpart. However, the impact of missing modality in different clients, also called modality incongruity, has been greatly overlooked. This paper, for the first time, analyses the impact of modality incongruity and reveals its connection with data heterogeneity across participating clients. We particularly inspect whether incongruent MMFL with unimodal and multimodal clients is more beneficial than unimodal FL. Furthermore, we examine three potential routes of addressing this issue. Firstly, we study the effectiveness of various self-attention mechanisms towards incongruity-agnostic information fusion in MMFL. Secondly, we introduce a modality imputation network (MIN) pre-trained in a multimodal client for modality translation in unimodal clients and investigate its potential towards mitigating the missing modality problem. Thirdly, we assess the capability of client-level and server-level regularization techniques towards mitigating modality incongruity effects. Experiments are conducted under several MMFL settings on two publicly available real-world datasets, MIMIC-CXR and Open-I, with Chest X-Ray and radiology reports. 

\end{abstract}

\section{Introduction}
\label{submission}
Multimodal Learning (MML) \cite{xu2023multimodal, bayoudh2021survey} has recently emerged as a pivotal area in machine learning research.
Different modalities represent diverse features that are sourced from diverse domains but describe similar subjects, offering both shared and complementary information. The essence of MML lies in combining predictive insights from these different modalities to enhance model performance. Despite the effectiveness of MML, many existing methods are constrained by their reliance on complete modalities, which is often scarce in practice, particularly when dealing with numerous modalities. 
Real-world multimodal data often presents inherent challenges with missing or incomplete modalities posing significant hurdles in the learning process \cite{aguilar2019multimodal,ma2021smil,jaques2017multimodal,pham2019found,parthasarathy2020training}. The presence of missing modalities within the multimodal datasets introduces complexities that traditional models struggle to accommodate, demanding specialized techniques to ensure effectiveness.

\begin{figure}[t]

    \centering
\includegraphics[width=1\columnwidth]{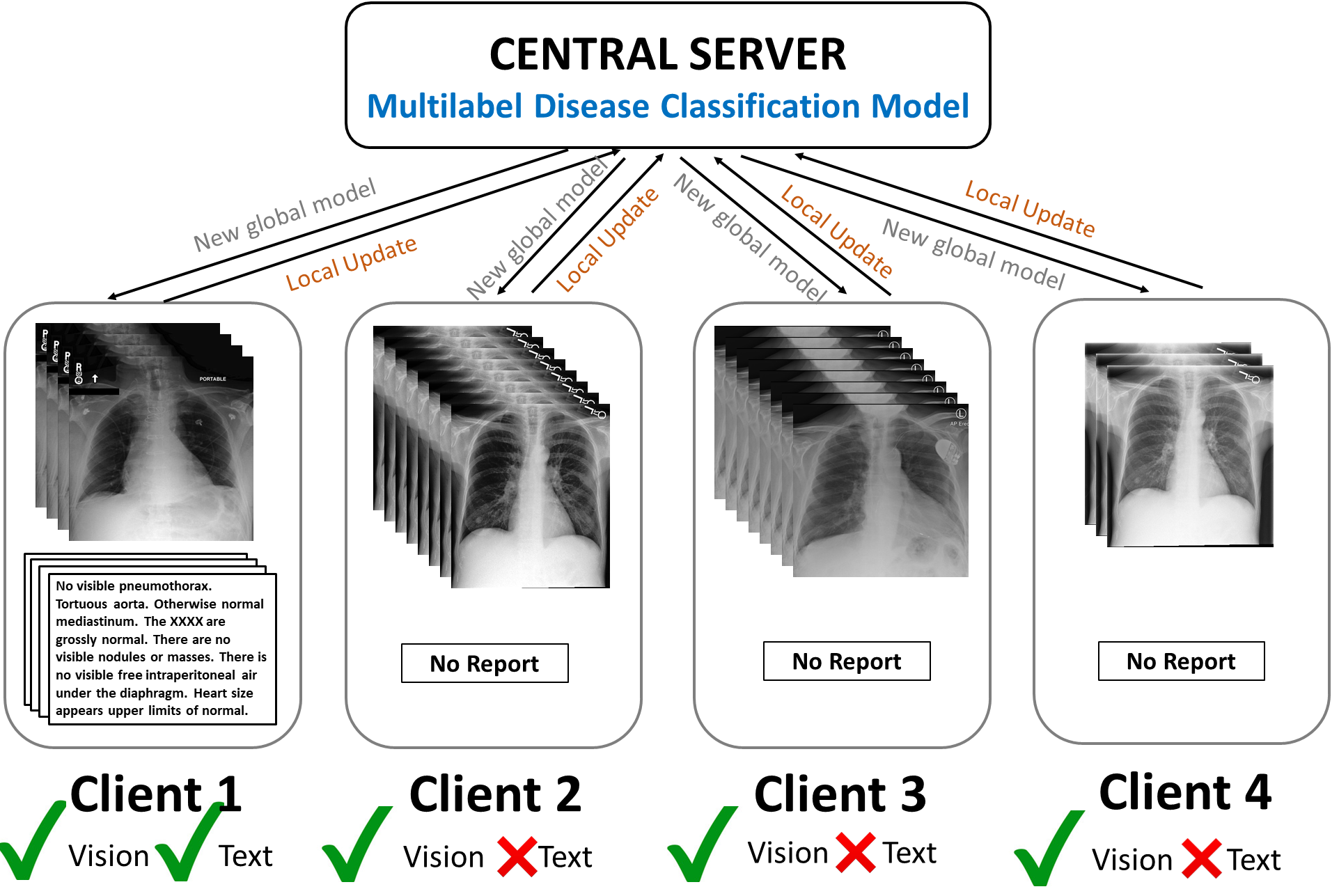}
    \caption{Overview of problem settings. Here, only 1 out of 4 clients have both modalities, \textit{i.e.},  CXR image and radiology report.} 
\label{fig:1}
\end{figure}

Typically, MML works focus on centralized training, requiring collection and storage of multimodal data on a server for training the models, leading to privacy concerns. The drawbacks of centralized learning has inspired researchers to develop and apply Federated Learning (FL) that enables various clients to collaboratively train models without sharing local data \cite{mammen2021federated,aledhari2020federated,li2021survey,zhu2021federated,huang2022learn}. Tackling modality incongruity is crucial in realistic Multimodal Federated Learning (MMFL) as presence of particular modalities across clients might vary, leading to poor performance. 

Most existing MMFL works \cite{xiong2022unified,agbley2021multimodal, salehi2022flash,qayyum2022collaborative, nandi2022federated, chen2022towards, wei2021multi} assume the presence of all modalities in each client. Despite being a critical question, investigation on the impact of missing modality \cite{le2023fedmekt,zhao2022multimodal,chen2022fedmsplit,yu2023multimodal} during training has been limited. Intuitively, multimodal models are expected to be more powerful than unimodal models. Therefore, it follows that multimodal clients involved in FL should have better performance than their unimodal versions owing to the availability of complementary information from additional modalities. However, it is not evident how the presence of both unimodal and multimodal clients impact the performance of MMFL in practice. This is particularly interesting as MML models are assumed to be more robust to missing modalities owing to possible redundancy between modalities. As a result, even if some clients are missing modalities, the other modalities should be able to compensate the loss. On the other hand, multimodal integration has been observed to be vulnerable to incomplete or missing modalities in centralized setting as MML models possess larger input dimension than unimodal models and the missing input dimensions may hamper the model training. To summarise, the impact of clients missing some modalities in MMFL is not well-known.

This naturally leads to the question: \textbf{Does an incongruent MFFL system benefit over unified FL by leveraging the extra modality present in multimodal clients?} Another related question is: Does the modality incongruity vary based on client heterogeneity? These questions are particularly crucial as addressing these questions can potentially help set up a practically beneficial MMFL system among clients in real-world scenarios. 
We strongly believe that this paper will facilitate decision making and provide easy, feasible solutions to alleviate the impact of modality incongruity in MMFL.



In this paper, we attempt to address these critical questions related to the absence of text modality in incongruent MMFL settings. However, we are aware that the investigation is task-specific and model architecture-sensitive. In other words, varying MMFL settings, target tasks, modalities, model architectures \textit{etc.}, can bias the results due to the presence of multiple variables influencing the learning outcome. Notably, there are many multimodal tasks and innumerable existing model architectures that could be explored in this context. However, the objective of this work is to primarily reveal different insightful aspects of MMFL by varying the presence of the primary modality in clients (text) instead of varying model architectures or proving its generalizability across a large number of MML tasks. In this work, we particularly choose a real-world multimodal problem using Medical Vision and Text (Report) modalities. We address a long-tailed multi-label disease classification problem (with 14 categories) from Chest X-Ray images and radiology reports. In this setting, some clients possess both images and radiology reports whereas the others possess only images as shown in Fig. 1. We include two publicly available datasets - MIMIC-CXR and NIH-Open-I in our study. Our primary contributions are:
\begin{enumerate}
    \item To the best of our knowledge, this is the first work that investigates modality incongruity effects in various heterogeneous MFFL settings. We empirically determine the conditions under which an incongruent MFFL system performs worse than the corresponding unimodal FL system in the context of non-IID data distribution. This reveals important considerations of designing a practical MFFL system with mixed unimodal and multimodal clients as well as suggests plausible modifications to improve performance.
    
    \item We first demonstrate how the variation of self-attention masks in the multimodal client(s) vary the effectiveness of information fusion in incongruent MFFL system.
    
    \item We transform the incongruent MFFL problem to pseudo-congruent MFFL by introducing a modality translation technique in unimodal clients and demonstrate its performance across varied MFFL settings as a direct way of mitigating modality incongruity.

    \item We introduce regularization schemes in unimodal and multimodal clients to achieve a client-invariant representation despite modality incongruity that includes the incorporation of proximal loss, contrastive loss, and modality-aware consistency regularization loss.

    \item We also demonstrate the potential of leveraging unlabeled data (both unimodal and multimodal) on the server to mitigate the modality incongruity issues via ensemble distillation and client model fine-tuning.
    
\end{enumerate}

\section{Related Works}
Multimodal learning leverages complementary information in multimodal data to enhance performance in various computer vision tasks \cite{zhang2021deep,moon2022multi}. A key focus in this field is multimodal fusion, which aims at effectively combining multimodal data. A prevalent technique, early fusion, merges different modalities through feature concatenation, a method extensively used in prior research \cite{poria2016convolutional, barnum2020benefits}. However, \cite{zadeh2017tensor} introduced a product operation for fusion, promoting greater interaction among modalities. Similarly, \cite{liu2018efficient} employed modality-specific factors for efficient low-rank fusion.
Another line of work maximized canonical correlations between two modalities, thereby uncovering their shared structure or common information \cite{andrew2013deep,kakade2007multi}. 
In terms of integrating supplementary information, a common approach involves using deep neural networks to derive abstract representations from each modality, which are then combined in various ways. For instance, \cite{mehrizi2018toward} adopted a straightforward concatenation approach for fusing different representations. Meanwhile, \cite{song2016multi} used both element-wise and weighted element-wise multiplication for modality fusion. However, all these works assume the presence of all modalities.

The current approaches to addressing missing modality in centralized setting can be categorized into three main types. The first category utilizes a data augmentation or drop-out strategy, as outlined by \cite{parthasarathy2020training}, which involves the random omission of input data to simulate missing modality scenarios. The second category employs generative techniques, as demonstrated by \cite{li2018video,cai2018deep,suo2019metric,lee2023unified,du2018semi}, which focus on synthesizing absent modalities using available modalities. The third category, highlighted by works such as \cite{aguilar2019multimodal,pham2019found,han2019implicit,wang2020transmodality}, concentrates on developing joint multimodal representations that encapsulate relevant information from the various modalities.

There have been various efforts to adapt FL in multimodal tasks. \cite{xiong2022unified} modified the FedAvg algorithm to suit a multimodal context. \cite{liu2020federated} implemented FL to utilize multiple datasets from diverse distributions, enhancing performance. 
\cite{yu2022multimodal} implemented contrastive learning to ensemble diverse local models in a federated system, based on their output representations. \cite{guo2023pfedprompt} adapted prompt training techniques to integrate large foundational models into FL systems, facilitating the connection between vision and language data. \cite{zhao2022multimodal} suggested giving greater weight to multimodal clients during aggregation. 
 A similar work, FedCMR \cite{zong2021fedcmr}, focused on the federated cross-modal retrieval task, addressing the challenge of bridging gaps in representation spaces through a weighted aggregation method that considered the local data volume and the number of categories. However, analysis of modality incongruity and its connection with data heterogeneity has not been discussed.

\section{Preliminaries and Problem Setup}
\textbf{Problem Formulation:} 
We consider a multilabel classification task within MMFL setting with $q$ multimodal and $n$ unimodal clients denoted as $\{ K^1, K^2,..., K^q\}$ and $\{K^{q+1},K^{q+2},..., K^{q+n} \}$ respectively. A sample datapoint in the dataset $D_m$ for a multimodal client $K_m$ along with its label(s) is denoted as $\{(X_i^m, Y^m)\}_{i=1}^{N_m}, m=1,2,...,q$, where $N_m$ denotes the number of modalities in $D_m$. The dataset $D_u$ for a unimodal client $K_u$ is denoted as $\{(X^u, Y^u)\}, u=q+1,..., q+n$. In this work, we only consider two modalities $(m=2)$ for multimodal clients (\textit{i.e.}, image and text) whereas only image for unimodal clients. Our goal is to minimize the following loss:
    $ \mathcal{L}(w)=\sum_{m=1}^q \mathbb{E}_{\{(X_i^m, Y^m)\}_{i=1}^{N_m}) \sim \mathcal{D}^m}\left[\mathcal{L}_m(w ;\{(X_i^m, Y^m)\}_{i=1}^{N_m})\right] + \sum_{u=1}^n \mathbb{E}_{\{(X^u, Y^u)\} \sim \mathcal{D}^u}\left[\mathcal{L}_u(w ;(X^u, Y^u))\right] $

\textbf{Datasets:}
We utilized the MIMIC-CXR \cite{johnson2019mimic} and NIH Open-I \cite{demner2016preparing} datasets. Although both MIMIC-CXR and Open-I consist of chest X-ray images and report pairs, the two datasets have different characteristics since they were collected from separate institutions and the diagnostic information represented by the two X-ray image sets are differently distributed (See Fig. 6 and 7 of Appendix $A$). The MIMIC-CXR dataset comprises 377,110 Chest X-ray images along with their corresponding free-text reports. 
Our experiments were conducted exclusively on 91,685 unique frontal view image-report pairs. These were divided according to the official MIMIC-CXR split (89,395 for training, 759 for validation, and 1,531 for testing). The other dataset, Open-I, includes 3,851 reports and 7,466 Chest X-ray images out of which 3,547 frontal view image-report pairs have been used.
There are 14 disease classes in the datasets, \textit{viz.}, Support Device, Pleural Effusion, Consolidation, Pneumothorax, Lung Opacity, Enlarged Cardiomediastinum, Atelectasis, Others, Cardiomegaly, Lung lesion, Edema, Fracture, Pneumonia, Pleural other and No finding. 
A mild imbalance was observed in MIMIC-CXR where the class ratios ranged from $13.39\%$ (support devices) to $1.2\%$ (pneumonia, and pleural other). On the other hand, a severe imbalance was observed in Open-I with the maximum class ratios of $28.8\%$ (Others, and cardiomegaly) and the minimum of $1.07\%$ (support devices) as shown in Appendix A.

\textbf{Federated Learning settings: }
We investigate the modality incongruity effects in both IID and non-IID settings. Following previous works \cite{chen2020fedbe,xiong2023feddm,saha2023rethinking,acar2021federated,li2021model,xiong2023feddm}, we use Dirichlet distributions with $\gamma = 100$ for simulating IID client data partition and $\gamma = 0.1, 0.5$ for non-IID partition. We evaluate the model performances with 4 clients under fully multimodal and unimodal settings. We confine our study to only 4 clients as in most cases this depicts a realistic number of collaborating institutions in healthcare. Besides, it is a deliberate methodological choice. Limiting the study to 4 clients allows for a more controlled and detailed analysis. With a higher number of clients, the complexity increases, potentially diluting the clarity and specificity of insights into the individual contributions and interactions of multimodal and unimodal clients. This focused approach ensures a more precise and meaningful understanding of the dynamics at play in such federated learning environments. For analyzing modality incongruent MMFL, we vary the ratio of multimodal and unimodal clients as 1:3 and 3:1. See Appendix $B$ for more details.

\textbf{Multimodal Learning settings and notations:}
For a given Chest X-Ray $v$, we denote the flattened visual feature from the last CNN layer as $v = \{ v_1, v_2,..., v_K\} $ and location feature as $l = \{l_1, l_2,...,l_K\} $ where $K$ denotes the number of visual features. The final visual embedding is $\Tilde{v}_i=v_i+l_i+s_V$ where $s_V$ is a semantic embedding vector for visual features. These features are projected into the final embedding space with same dimension as language embedding space via a fully connected layer. For the corresponding report, the text embedding is denoted as $w=\{ w_1, w_2,...,w_N\}$ and the corresponding positional embedding as $p= \{p_1, p_2,...,p_N\}$. The final language features are expressed as $ \Tilde{w}= w_i+p_i+s_L$, where $s_L$ is semantic embedding vector for language features. The visual and language embeddings are concatenated to form joint embedding for feeding into the multimodal transformer in multimodal clients as $\Tilde{J}= \{{S}, v_1, v_2,..., v_K, \Bar{SEP}, w_1, w_2,...,w_N, {E} \}$ where the embedding length $L_{emb}= N+K+3$. Here, we obtain the start, separation and end tokens ${S}, \Bar{SEP}, {E}$ by adding the special tokens with corresponding position and semantic embedding. For unimodal clients, we apply padding for the missing text embeddings. We learn unified, contextualized representation of CXR and reports using single BERT-based transformer encoder model \cite{kenton2019bert,moon2022multi} and attach 14 linear heads to the transformer to address the 14-class multilabel classification model. For more implementation details, see Appendix $B$.

\section{Modality incongruity in MFFL}
We start by defining how modality incongruity can be quantified in MFFL. We particularly explore three different settings: (a) a fully multimodal setting where all clients have multimodal data - both Chest X-Ray (CXR) images and radiology reports, (b) a fully unimodal setting where all clients have unimodal data, \textit{i.e.}, only CXR, and (c) a mixed unimodal-multimodal setting where some clients have both CXR and reports while others only possess CXR. We further vary the last setting by varying the proportion of unimodal and multimodal clients. We evaluate modality incongruity by comparing model performance in each setting. The higher the performance difference between (a) and (c), the more the modality incongruity effects. If (c) performs poorer than (b), we consider it to be severely modality incongruen. With this background, we first ask the question:

\begin{table}
    \centering
    \caption{MMFL Performance with varying degree of 
 heterogeneity. M and U denotes the number of multimodal and unimodal clients.} 
    \scalebox{0.7}{
    \begin{tabular}{|l|llll|llll|}
    \hline 
       \textbf{Data} &  \multicolumn{2}{|c|}{\textbf{Open-I}} & \multicolumn{2}{|c|}{\textbf{MIMIC CXR}} &  \multicolumn{2}{|c|}{\textbf{Open-I}} & \multicolumn{2}{|c|}{\textbf{MIMIC CXR}} \\ \hline
        \textbf{Partition} & \textbf{AUC} & \textbf{F1} & \textbf{AUC} & \textbf{F1} & \textbf{AUC} & \textbf{F1} & \textbf{AUC} & \textbf{F1} \\ \hline\hline
     
      & \multicolumn{4}{|c|}{ \textbf{M:U = 3:1}} &  \multicolumn{4}{|c|}{ \textbf{M:U = 1:3}}\\ \hline
        IID & 77.64 & 29.35 &  96.43 & 81.90 & 67.61 &  20.43 &  94.51 & 80.85 \\  
       $\gamma$= 0.5 & 74.27 &  26.60& 87.92 &  77.28 & 67.33 &  20.36 &   80.01 & 72.75  \\ 
         $\gamma$= 0.1  &58.56 &  25.38 &76.84 & 69.67 & 53.69 &  22.38 &  70.14 & 66.86  \\ \hline
          &  \multicolumn{4}{|c|}{ \textbf{Fully multimodal (M:U = 4:0)}}  &  \multicolumn{4}{|c|}{ \textbf{Fully unimodal (M:U = 0:4)}}\\ \hline
        IID & 93.01 &  46.54 & 98.00 & 84.63 & 66.85  & 19.76 &  94.05  &  80.70 \\ 
       $\gamma$= 0.5  & 88.79 &  39.84& 96.89 &  83.37 & 76.12 &  29.77 & 92.26  & 79.85  \\ 
       $\gamma$= 0.1  & 84.37 &  42.12& 96.10  & 82.55 & 75.18 &  36.75 & 91.12 &  78.80 \\ \hline

    \end{tabular}}
\end{table}

\textbf{Question:} \color{blue}Is incongruent MFFL more beneficial than unimodal FL?\color{black}

\textbf{Observation:} \color{blue} We empirically conclude that incongruent MFFL outperforms its unimodal version only under homogeneous setting, \textit{i.e.}, IID data partition. The unimodal FL performance surpasses that of incongruent MFFL for both multimodal client proportions under heterogeneous or non-IID data distribution across clients. \color{black}

First, we empirically validate that the model performance of (b) improves over (a) under homogeneous setting with Dirichlet coefficient $\gamma=100$ in Table 1. As observed, when 3  \color{cyan}(or 1) \color{black}out of 4 clients are multimodal clients in IID settings, the model AUC drops respectively by $15.37$ and $1.57$ \color{cyan}(or $25.40$ and $3.49$) \color{black} below fully multimodal FL settings in Open-I and MIMIC respectively. However, the model performance is better than unimodal FL performance in all the above cases. 
However, we observe that under non-IID partitoning with Dirichlet coefficient $\gamma=0.1, ~0.5$, the MMFL performance severely deteriorates and is even worse than the unimodal settings for both the datasets. For $\gamma=0.1$, when only 3 \color{cyan}(or 1) \color{black}out of 4 clients possess both CXR and report, AUC drops respectively by $16.62$ and $14.28$ \color{cyan}(or $21.49$ and $20.98$) \color{black}below fully unimodal settings in Open-I and MIMIC respectively. This indicates that the presence of unimodal clients adversely impacts the multimodal ones in mixed, heterogeneous MFFL which results in sub-optimal utilization of the reports even where they are available. It is observed that the impact of modality incongruity increases with increase in heterogeneity. Besides, even though the model performance decreases with decreasing proportion of multimodal clients, the degradation is relatively low. The replacement of first multimodal client by unimodal client decreases the performance by $25.81$ and $19.26$, whereas the replacement of two more clients only drops the performance by another $4.87$ and $6.70$ in Open-I and MIMIC respectively.


\section{Self-attention Mechanisms}
\begin{figure}

    \centering
\includegraphics[width=1.02\columnwidth]{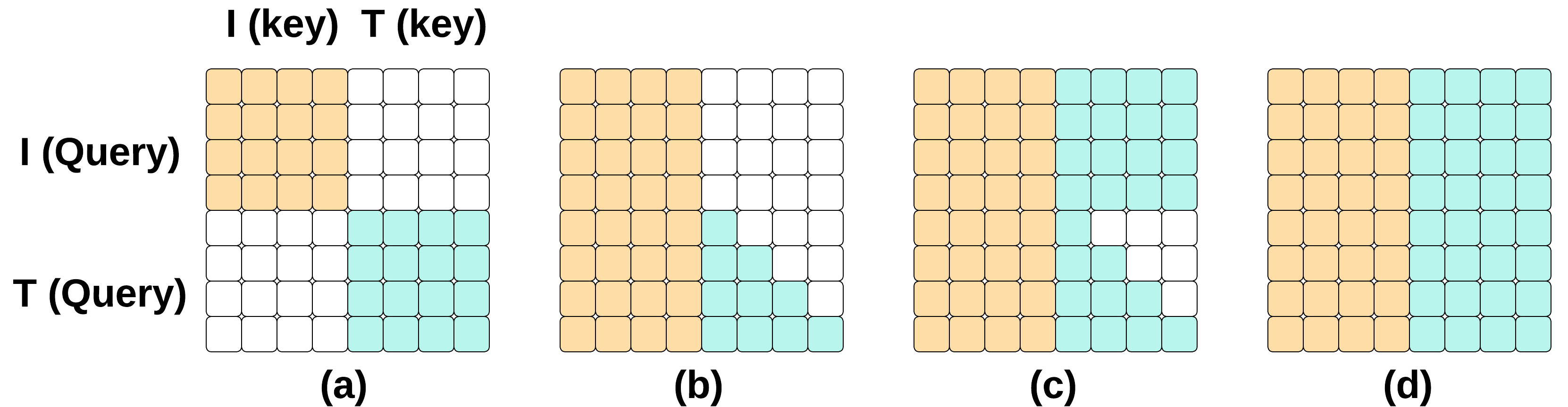}

    \caption{Four self-attention schemes used in multimodal client(s). (a) Isolated (b) Causal (c) Partially Bidirectional (d) Bidirectional.  } 
    
\label{fig:2}
\end{figure}

We use isolated self-attention mask as baseline that restricts the interaction between two modalities in multimodal clients as shown in Fig. \ref{fig:2} (a).
In this section, we further investigate three other self-attention mechanisms to potentially facilitate the model's learning of multi-modal representation that is more robust to the adverse influence of unimodal clients in incongruent MFFL. 
Each of these masks offers a unique way of handling the interactions between image and text modalities, which is crucial in our study of modality incongruity in MFFL. 
By experimenting with these different masks, we gain insights into how different levels and types of modality integration impact the learning process, especially in the presence of unimodal clients.

The self-attention mask $M \in R^{L_{emb} \times L_{emb}}$ is denoted as:
\begin{equation}
\small
M_{j k}=\left\{\begin{array}{l}
0, \quad(\text { attention allowed }) \\
-\infty, \quad(\text { attention not allowed })
\end{array} \quad j, k=1, \ldots, L_{emb} .\right.
\end{equation}
In the self attention module, each attention head can be represented as: $\text { \textit{Attention} }=\operatorname{softmax}(S A+M) V, \quad S A=\frac{Q K^T}{\sqrt{d_k}}$
where $Q, K, V, d_k$ respectively indicates queries, keys, values and dimension of keys. 
Based on modality type, self attention matrix (SA) can be expressed in terms of four subparts: $
S A_{q, k}  =S A_{S_q: S E P_{q}, S_k: S E P_{k}} 
+S A_{S_q: S E P_{q}, W_{1 k}: E_{L k}} 
+S A_{W_{1 q}: S E P_{q}, S_k: S E P_{k} }
+S A_{W_{1 q}: E_{q}, W_{1 k}: E_{k}}
$. Below we discuss four types of self-attention and justification behind their usage in the work. For more details, see Appendix C.

\textbf{Isolated Self-Attention:} It serves as a baseline and is crucial for understanding the model performance in a scenario where the two modalities are treated independently. By restricting the interaction between image and text modalities in multimodal clients, we can assess the inherent capabilities of each modality to contribute to the learning process. This is particularly important in our context, where some clients only have one modality (CXR), and we need to understand how much each modality can contribute on its own. 


\textbf{Causal Self-Attention:} The causal mask introduces a controlled interaction between the modalities. It allows language features to attend to both preceding words and visual features, but prevents visual features from attending to language features. This asymmetrical attention is particularly useful in scenarios where the temporal or sequential nature of one modality might inform the understanding of the other, but not vice versa. This is especially relevant in our case as this restricts the image embeddings to attend to the text which is missing in other modalities while still allowing the text to be guided by image in multimodal clients.



\textbf{Bidirectional Self-Attention:} By allowing unrestricted interaction between the image and text modalities, the bidirectional mask facilitates comprehensive context learning. 
This is essential for exploring the full potential of multimodal learning, especially in cases where the integration of modalities can lead to a more holistic understanding than either modality alone. This mask is particularly useful for scenarios where the interplay between text and image is complex and deeply intertwined. 

\textbf{Partially Bidirectional Self-Attention}: This aims to combine the benefits of both bidirectional and causal masks. It allows for the integration of image features with language features (like the bidirectional mask) while preserving the causal nature of language (like the causal mask). It is particularly beneficial in the current scenario if we want to leverage the rich context provided by the bidirectional approach but still need to maintain the sequential integrity of the text. It represents a middle ground, offering a balance between context richness and controlled information flow. 

\begin{table}
    \centering
    \caption{MMFL Performance with varying self-attention schemes} 
    \scalebox{0.65}{
    \begin{tabular}{|l|llll|llll|llll|llll|}
    \hline 
     &  \multicolumn{4}{|c|}{ \textbf{$\gamma=0.5$}}&  \multicolumn{4}{|c|}{$\gamma=0.1$}\\ \hline
       \textbf{Self} &  \multicolumn{2}{|c|}{\textbf{Open-I}} & \multicolumn{2}{|c|}{\textbf{MIMIC CXR}}&  \multicolumn{2}{|c|}{\textbf{Open-I}} & \multicolumn{2}{|c|}{\textbf{MIMIC CXR}} \\ \hline
        \textbf{Attention} & \textbf{AUC} & \textbf{F1} & \textbf{AUC}  & \textbf{F1 score}& \textbf{AUC}  & \textbf{F1 score}& \textbf{AUC}  & \textbf{F1} \\ \hline\hline
     
      & \multicolumn{8}{|c|}{ \textbf{M:U = 1:3}}\\ \hline
        Isolated & 67.33 &  20.36 & 80.01 &  72.75 &53.69 & 22.38& 70.14 & 66.86 \\ 
      Causal  & 68.11 &  25.55 & 82.89 &  73.85 & 54.05 &  22.09 & 72.50 & 67.91\\ 
       parBi  & 70.71 &  \textbf{25.77} & 84.75 &  76.50 & 54.46 & \textbf{22.66} & 75.36 &  68.60 \\
       Bi  & \textbf{70.79} & \textbf{25.77} &\textbf{85.66} &  \textbf{77.07} & \textbf{57.55} &  19.75 & \textbf{76.09} &  \textbf{68.63}\\ 
      \hline
     &  \multicolumn{8}{|c|}{ \textbf{ M:U = 3:1}}\\ \hline
        Isolated & 74.27 & 26.60 & 87.92 &  77.28 & 58.56  & 25.38  & 76.84 &  69.67\\ 
      Causal  & 74.48 &  26.97 & 88.89 &  78.32 & 58.86 & \textbf{28.06} & 78.73 &  71.08\\ 
      parBi  & 75.38  & 27.77 & 90.05 &  79.88 & 59.43 &  26.30  & 80.10 &  \textbf{72.75} \\
       Bi  & \textbf{76.76}  & \textbf{30.99} & \textbf{90.20} & \textbf{80.76} & \textbf{61.84} &  26.11 & \textbf{81.13} & 72.42\\ 
        
      \hline

    \end{tabular}}
\end{table}

\textbf{Performance analysis:} The performance of various self-attention schemes in MMFL is summarized in Table 2. As shown in the table, all the other masks improve the performance over isolated masks. Overall, Bidirectional self-attention mask shows the best performance and outperforms the isolated mask by around $3.27\%$ and $4.54\%$ for Open-I and MIMIC respectively in terms of AUC score. The improvement is relatively higher with higher heterogeneity in data partition and for lesser proportion of multimodal clients. However, while variation of self-attention masks show slight improvement in performance, it fails to enable MMFL to surpass the corresponding unimodal performance. This demonstrates that \color{blue}varying self-attention masks can only act as an assisting agent to boost the MMFL performance but not as a stand-alone factor towards achieving better performance than unimodal FL in heterogeneous settings. \color{black}

\section{Incongruent to pseudo-congruent MFFL}
\begin{figure}[t]
    \centering
\includegraphics[width=0.8\columnwidth]{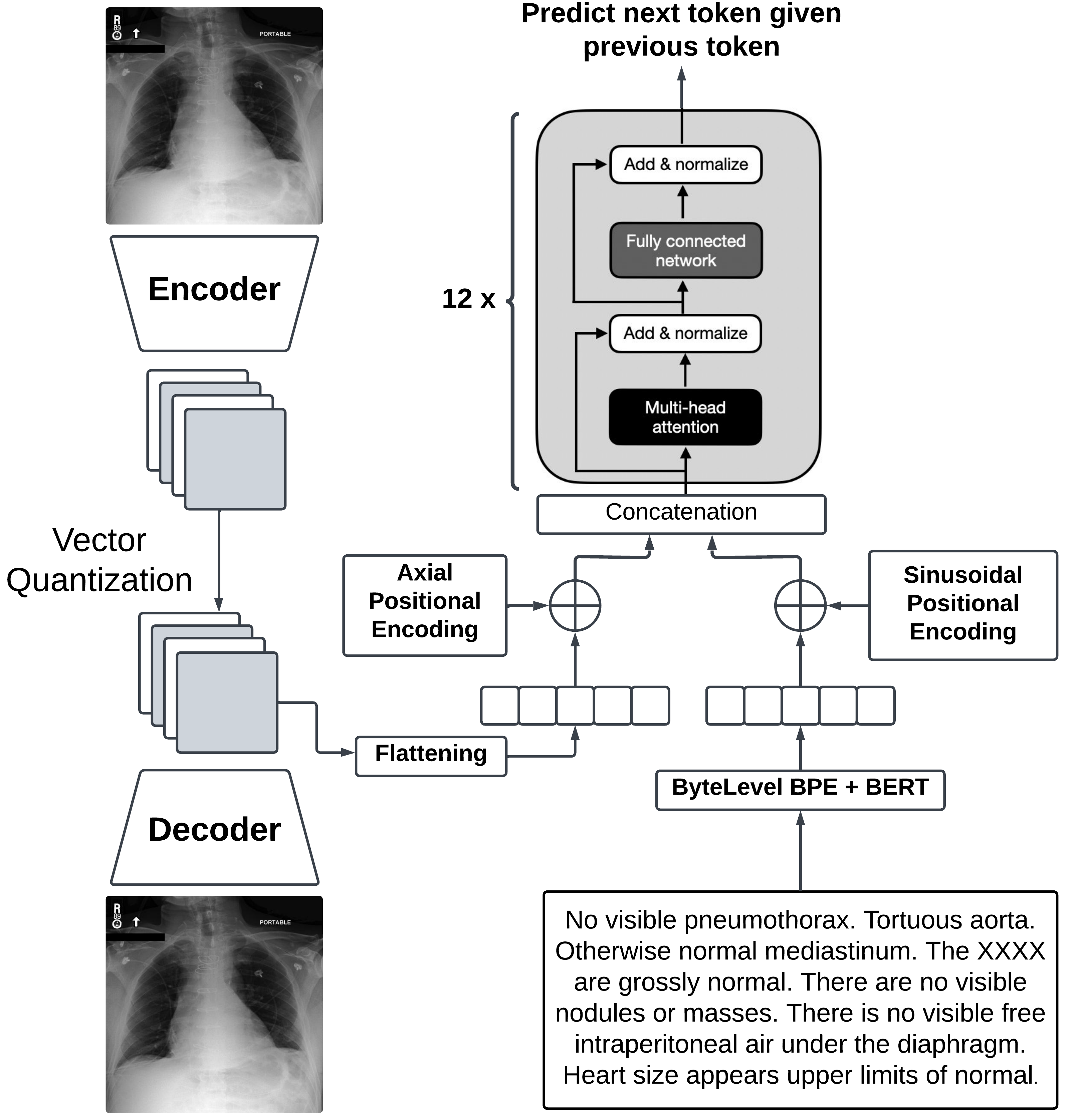}
    \caption{Modality Imputation Network (MIN) Training procedure} 
\label{fig:3}
\end{figure}

\textbf{Modality Imputation Network:}
We convert the incongruent MFFL setting to pseudo-congruent MFFL by introducing a Modality Imputation Network (MIN) to generate radiology reports based on CXR in the unimodal clients as shown in Fig. \ref{fig:3}. This imputation is performed prior to the start of federated learning procedure and does not add any computational overhead. For this, we first utilize VQ-GAN as the image tokenizer, which is composed of an encoder, a decoder, and a learnable codebook of fixed size. The encoder first transforms the CXR image $\mathbf{x} \in \mathbb{R}^{H \times W \times 3}$ into a continuous feature space $\mathbf{z} \in \mathbb{R}^{h \times w \times d_z}$. Consequently, it is quantized into a series of discrete tokens $\{ v_1,v_2,...,v_{h \times w}\}$ by identifying the nearest code embedding in the codebook through nearest neighbor search. The decoder reconstructs the original input from these discrete codes. This approach enables the model to develop a concise and discrete representation of the images. Next, we split each report in the multimodal client into individual word tokens using a byte-level BPE tokenizer, and encase these tokens with specific markers. The ultimate embeddings for the report are derived by combining the word embeddings with a sinusoidal positional embeddings as shown in Fig. 3.

We introduce a BERT-based cross-modal Transformer architecture \cite{kenton2019bert,lee2023unified} and train our model with a causal attention mask in the multimodal client which allows the model to learn about the radiological report in a sequence, conditioned on the CXR images. In order to efficiently manage long-range sequences under limited computational resources, we employ an efficient attention mechanism called Performer \cite{choromanski2020rethinking}. 
During training in a multimodal client, we concatenate CXR and report embeddings from the same subject as depicted in Fig. \ref{fig:3} and feed it into the model. The problem is considered to be a sequence generation task and model is trained to minimize the negative log-likelihood of predicting the next token based on the preceding tokens. The loss function is: $
L= \sum_{i=1}^n-\log P\left(w_i \mid w_{0: i-1}\right)+\sum_{i=1}^m-\log P\left(v_i \mid w, v_{0: i-1}\right) $
where $n$ = {text sequence length} + 2 and $m= h \times w + 2$ as $w_0, w_n, v_0, v_m$ are special tokens.
See Appendix $B$ for more details. After the training procedure is completed, we freeze the pre-trained model and use it for generating reports in unimodal clients thereby transforming incongruent MFFL to semi-congruent or pseudo-congruent MFFL. 

\begin{table}
    \centering
    \caption{MMFL Performance with MIN} 
    \scalebox{0.68}{
    \begin{tabular}{|l|llll|llll|}
    \hline 
       \textbf{Data} & \multicolumn{4}{|c|}{\textbf{Open-I}} &  \multicolumn{4}{|c|}{\textbf{MIMIC CXR}}  \\ \hline
        \textbf{Partition} &\multicolumn{4}{|c|}{\textbf{BLEU-4}} &\multicolumn{4}{|c|}{\textbf{BLEU-4}}\\ \hline\hline

   &C1 (T)&C2&C3&C4&C1 (T)&C2&C3&C4 \\ \hline
       
       $\gamma$= 0.5 & 0.051 & 0.048 & 0.046 & 0.046 & 0.067 & 0.064 & 0.061 & 0.061 \\ 
         $\gamma$= 0.1  & 0.048 & 0.043 & 0.040 & 0.041 & 0.061 & 0.052 & 0.054 & 0.054   \\ \hline \hline
         &\textbf{AUC}&\textbf{Recall}&\textbf{Prec}&\textbf{F1}&\textbf{AUC}&\textbf{Recall}&\textbf{Prec}&\textbf{F1} \\ \hline
        & \multicolumn{8}{|c|}{ \textbf{M:U = 1:3}}\\\hline
       $\gamma$= 0.5 & 78.42 & 28.74 & 54.19 & 37.56 & 92.86 & 78.16 & 83.33 & 81.23 \\ 
         $\gamma$= 0.1  & 76.78 & 46.26 & 30.94 & 37.08 & 92.08 & 82.96 & 78.64 & 80.74   \\ \hline
          & \multicolumn{8}{|c|}{ \textbf{M:U = 3:1}}\\\hline
          $\gamma$= 0.5 & 81.24 & 86.68 & 29.49 & 44.01 & 93.45  & 82.31 & 84.74  & 83.68 \\ 
         $\gamma$= 0.1  & 79.61 & 32.48 & 50.00 & 38.67 & 93.36 & 74.25 & 90.09 & 81.30\\ \hline

    \end{tabular}}
\end{table}

\begin{figure*}[t]
    \centering
\includegraphics[width=2.10\columnwidth]{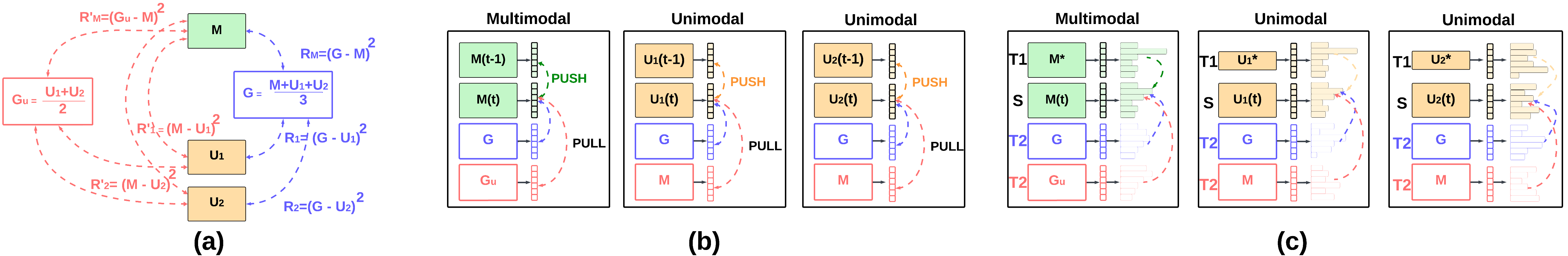}
    \caption{Illustration of client-level solutions in a 3-client FL scenario - one multimodal client ($M$) and two unimodal clients ($U_1$ and $U_2$). (a) shows the model-based regularization technique of FedProx (in blue) and FedMultiProx (in red). The global model $G$ is replaced by $G_u$ in multimodal clients and $M$ in unimodal clients. (b) shows the representation-based regularization technique of MOON (in blue) and MultiMOON (in red). (c) shows the Modality-aware Knowledge Distillation technique (MAD) and MAD+. $M^*, U_1^*, U_2^*$ represent pre-trained models, \textit{i.e.}, the first teacher model $T_1$. $G$ denotes the second teacher model $T_2$ in all the clients for MAD. For MAD+, $G_u$ denotes the second teacher model in the multimodal client and $M$ denotes the second teacher model in the unimodal clients.}  
\label{fig:4}
\end{figure*}
\begin{table*}[t]
    \centering
    \caption{MMFL Performance with client- and server-level solutions. T and I indicate the presence of Text and Image respectively} 
    \scalebox{0.76}{
    \begin{tabular}{|l|ll|ll|ll|ll|ll|ll|ll|ll|}
    \hline 
     &  \multicolumn{8}{|c|}{ \textbf{$\gamma=0.5$}}&  \multicolumn{8}{|c|}{$\gamma=0.1$}\\ \hline
     
       \textbf{Methods} &  \multicolumn{2}{|c|}{\textbf{Open-I}} & \multicolumn{2}{|c|}{\textbf{MIMIC CXR}}&  \multicolumn{2}{|c|}{\textbf{Open-I}} & \multicolumn{2}{|c|}{\textbf{MIMIC CXR}}  &  \multicolumn{2}{|c|}{\textbf{Open-I}} & \multicolumn{2}{|c|}{\textbf{MIMIC CXR}}&  \multicolumn{2}{|c|}{\textbf{Open-I}} & \multicolumn{2}{|c|}{\textbf{MIMIC CXR}} \\ \hline
        \textbf{Methods} & \textbf{AUC} & \textbf{F1} & \textbf{AUC} & \textbf{F1}& \textbf{AUC} & \textbf{F1}& \textbf{AUC} & \textbf{F1} & \textbf{AUC} & \textbf{F1} & \textbf{AUC} & \textbf{F1}& \textbf{AUC} & \textbf{F1}& \textbf{AUC} & \textbf{F1} \\ \hline\hline
  
      & \multicolumn{4}{|c|}{ \textbf{M:U = 1:3}}  & \multicolumn{4}{|c|}{ \textbf{M:U = 3:1}} & \multicolumn{4}{|c|}{ \textbf{M:U = 1:3}}  & \multicolumn{4}{|c|}{ \textbf{M:U = 3:1}}\\ \hline
         FedAvg & 67.33 & 20.36 &80.01 &  72.75 & 74.27 & 26.60 & 87.92 &  77.28& 53.69 & 22.38 &70.14 & 66.86 & 58.56& 25.38   & 76.84 & 69.67 \\ \hline
         &  \multicolumn{16}{|c|}{ \textbf{Client-level solutions}}\\ \hline
       
        FedProx & 69.26 & 24.40 & 83.44 & 73.28 & 74.90 & 27.88 & 88.67 & 80.16& 56.37 & 23.85 & 71.08 & 69.15 & 60.20& 24.73 & 76.96 & 71.32 \\ 
      FedMultiProx  & 70.92 & 24.91 & 85.67 & 75.06 & 76.24 & 28.05 & 90.03 & 79.48 & 58.12 & 24.51 & 72.34 & 68.26 & 62.27 & 28.91 & 78.20 & 71.68\\ 
       MOON  & 68.29 & 22.45 & 82.00 & 73.98 & 74.75 & 28.82 & 85.73 & 75.24 &  55.64& 24.04 & 74.05 & 70.88 & 60.48 & 25.05 & 77.54 & 73.27 \\
        MultiMOON & 70.38 & 24.73 & 83.98 & 72.41 & 76.02 & 29.64 & 88.98 & 77.05 & 57.92 & 25.59 & 76.69 & 70.61 & 61.96 & \textbf{29.38} & 80.18 & 74.47\\ 
         MAD & 73.82 & \textbf{26.95} & 84.69 & 75.20 & 78.28 & 29.36 & 89.90 & 76.87 & 60.77 & 26.80 & 77.23 & 72.20 & 64.70 & 28.84 & 81.87 & 74.43 \\
        MAD+  & \textbf{74.39} & 26.74 & \textbf{86.92} & \textbf{78.65} & \textbf{79.05} & \textbf{29.96} & \textbf{91.34} & \textbf{80.23} & \textbf{61.80} & \textbf{27.49} & \textbf{80.83} & \textbf{73.18} & \textbf{66.62} & {29.18} & \textbf{84.43} & \textbf{75.02} \\ 
      \hline
         &  \multicolumn{16}{|c|}{ \textbf{Server-level solutions} (utilizing additional data)} \\ \hline
            
       FedDF (I)  & 74.03 & 29.36 & 84.88 & 74.82 & 78.29 &  32.03 & 88.20 & 78.90& 61.28 & 27.84 & 77.18 & 70.58 & 65.46 & 28.74 & 82.06 & 73.85 \\
        FedDF (T)  & 71.49 & 26.00 & 82.09 & 72.49 & 77.80 & 30.65 & 85.81 & 77.01 & 59.49 & 26.77 & 73.65 & 71.24 & 63.04 & 27.99 & 80.43 & 74.89 \\
       FedDF (I+T) & 76.72 & 33.94 & 85.13 & 75.45 & 80.22 & 35.18 & 90.82 & 79.90 & 63.93 & \textbf{29.80} & 80.01 & 74.16 & 68.10 & 30.30 & 86.50 & 76.78 \\ 
         
        LOOT (I)  & 75.06 & 34.98 & 86.85 & \textbf{78.19} & 79.94 & 33.74 & 90.33 & 81.28 & 62.49 & 28.02 & 82.22 & 74.54 & 66.02 & 28.85 & 86.08 & 75.47 \\
        LOOT (T)  & 73.84 & 27.43 & 83.10 & 71.36 & 78.55 & 31.67 & 89.39 & 79.88 & 60.36 & 27.88 & 78.10 & 71.93 & 65.38 & 28.33 & 83.14 & 74.30 \\
         LOOT (I+T)  & \textbf{79.36} & \textbf{38.73} & \textbf{89.97}  & 77.85 & \textbf{83.75} & \textbf{40.30} & \textbf{92.47} & \textbf{83.34} & \textbf{65.25} & {29.32} & \textbf{84.29} & \textbf{75.17} & \textbf{70.94} & \textbf{30.65} & \textbf{89.60} & \textbf{77.09} \\
      \hline

    \end{tabular}}
\end{table*}

\textbf{Performance analysis:}
Table 3 shows the report generation performance of MIN across all clients in terms of BLEU-4 score. As the model is trained on Client 1 (C1), we first validate its performance on the test set of the same client (C1(T)). For the other clients (C2-C4), we test the report generation performance on all local data samples. The mean BLEU-4 scores for Open-I \color{cyan}(MIMIC) \color{black} with $\gamma=0.1$ and $\gamma=0.5$ are  0.043 \color{cyan}(0.055) \color{black}and 0.048 \color{cyan}(0.063) \color{black}respectively. It is also observed from Table 3 that \color{blue}MIN enables the incongruent MFFL system to be more beneficial than unimodal FL in almost all cases. \color{black} \textit{Eg:} For downstream classification task with $\gamma=0.1$, incongruent MMFL with 1 and 3 multimodal clients surpass the respective unimodal FL AUC by $1.6$ and $4.43$ for Open-I and by $0.96$ and $2.24$ for MIMIC.
\section{Towards modality-invariance in MFFL}

The heterogeneous data distribution and modality incongruity lead to distributional modality gaps between the unimodal and multimodal clients thereby posing significant challenges. In this section, we intend to learn modality- and client-invariant representations to aid the information fusion process by bridging the gap between clients. In FL context, this can be achieved either from the client side by carefully constraining the clients or from the server side by leveraging some unlabeled publicly available dataset to learn generalizable representation despite modality shift.

\subsection{Client-level solutions}
Overall, we consider three primary ways of constraining or regularizing the clients to learn modality-invariant representations. In each of the following techniques, we improve upon the naive unimodal federated learning strategy by incorporating prior knowledge regarding the presence of particular modalities in different clients as shown in Fig.4.

\textbf{Model parameter-based regularization:} This approach incorporates a regularization term to effectively mitigate the influence of varying local updates, as in FedProx \cite{li2020federated}. 
Motivated by this, we introduce \textbf{FedMultiProx} in this work to reduce the model diversity among unimodal and multimodal clients originating from the variation of information content from client to client. Rather than constraining each client model to be more aligned with the global model, we specifically regularize the models in unimodal client groups to match the averaged model from multimodal client group and vice versa. This forces the model to focus particularly on modality incongruity effects by penalizing large deviations between the unimodal and multimodal client(s), thereby effectively keeping the local updates in these client groups closer to each other. Accordingly, the optimization objective in $m^{th}$ multimodal client is denoted as $ \min _{\theta_t^m} \mathbb{E} _{\{(X_i^m, Y^m)\}_{i=1}^{N_m}\sim \mathcal{D}^m}[\mathcal{L}_{CE}^m (\theta_t^m;\{(X_i^m, Y^m)\}_{i=1}^{N_m})+$ $\lambda\left\|\theta_t^m-\frac{1}{n}\sum_{u=1}^n \theta_{t-1}^u\right\|^2]$, where $\theta$ denotes network parameter. The objective in $u^{th}$ unimodal client can be denoted as $\min _{\theta_t^u} \mathbb{E}_{\{(X_1^u, Y^u)\}) \sim \mathcal{D}^u}[\mathcal{L}_{CE}^u (\theta_t^u;\{(X_1^u, Y^u)\})+\lambda\left\|\theta_t^u-\frac{1}{q}\sum_{m=1}^q \theta_{t-1}^m \right\|^2]$. $\lambda$ is a tuning hyperparameter.

\textbf{Representation-based regularization:} Another way of ensuring that the local updates are closely aligned with the representations learned by the global model is applying contrastive learning at the representation level or embedding space, thereby comparing and contrasting the feature representations derived from different models as in MOON \cite{li2021model}. Since the global model is expected to yield modality heterogeneity-agnostic representations, the objective is to minimize the disparity between the client representation ($z_t^k$) and global representation ($z_{t-1}^G$), while simultaneously maximizing the disparity between the client representation at current step ($z_t^k$) and previous step ($z_{t-1}^k$).
In this work, we propose \textbf{MultiMOON} by replacing the global model of MOON by averaged multimodal client model in unimodal client group and averaged unimodal client model in multimodal client group. For this, we individually replace $z^G_{t-1}$ by $z^M_{t-1}$ for unimodal clients and by $z^U_{t-1}$ for multimodal clients that enforces a stronger constraint that essentially bridges the modality heterogeneity gap between unimodal and multimodal clients. \textit{Eg:} The loss in a unimodal client can be denoted as: $\mathcal{L}^k_{\text {con}}(\theta_t^k;\theta_{t-1}^k;\theta_{t-1}^G;\{X_i^k\}_{i=1}^{N_k}) =-\log \frac{\exp \left(\operatorname{sim}\left(z_t^k, z_{t-1}^M\right) / \tau\right)}{\exp \left(\operatorname{sim}\left(z_t^k, z_{t-1}^{M}\right) / \tau\right)+\exp \left(\operatorname{sim}\left(z^k, z^k_{{t-1 }}\right) / \tau\right)}$
where $sim(\cdot,\cdot)$ and $\tau$ denote cosine similarity function and temperature parameter respectively.


\textbf{Consistency regularization:} The inter-client modality gap can also be addressed by applying consistency regularization at the logit level via knowledge distillation. To this end, we propose \textbf{Modality-Aware knowledge Distillation (MAD)} exploiting the global and local knowledge. We introduce a dual teacher model with the global model as one teacher and a frozen local model pre-trained solely on the local client data (unimodal or multimodal) as the other. The student network is trained via guidance from the logit outputs of both the teacher models, thereby indirectly reducing the gap between unimodal and multimodal feature representations in a given client. For this, we minimize the KL divergence of the student logits with respect to the logits of both the teacher models denoted as $\mathcal{L}_{MAD}^{k}=\sigma\left(z_{pre}^{k}\right) \log \frac{\sigma\left(z_{pre}^{k}\right)}{\sigma\left(z_{t}^{k}\right)} + \sigma\left(z_{t-1}^{G}\right) \log \frac{\sigma\left(z_{t-1}^{G}\right)}{\sigma\left(z_{t}^{k}\right)}$ where $z_{pre}$ denotes the locally pretrained model embedding and $\sigma$ denotes softmax function.
Next, following our previous modifications, we propose a variant of MAD, which we term \textbf{MAD+}, by replacing the global model with averaged multimodal (or unimodal) client model for unimodal (or multimodal) client groups. Employing knowledge distillation under this setting forces the training to focus on effective balancing of distilled knowledge between unimodal and multimodal clients thereby achieving better modality invariance. The loss function is:  $\mathcal{L}_{MAD+}^{k}=\sigma\left(z_{pre}^{k}\right) \log \frac{\sigma\left(z_{pre}^{k}\right)}{\sigma\left(z_{t}^{k}\right)} + I\{k=u\} \left[\sigma\left(z_{t-1}^{M}\right) \log \frac{\sigma\left(z_{t-1}^{M}\right)}{\sigma\left(z_{t}^{k}\right)} \right]  +I\{k=m\}\left[\sigma\left(z_{t-1}^{U}\right) \log \frac{\sigma\left(z_{t-1}^{U}\right)}{\sigma\left(z_{t}^{k}\right)}\right]$. I is indicator function.


\subsection{Server-level solutions}
We investigate whether the presence of some unlabeled data on server can help us to reduce the modality gap between unimodal and multimodal client models. For this, we particularly consider three different modality settings (only CXR, only report, and both modalities) each for two datasets - with and without domain gap with respect to the client data in the server. For the latter, we utilize a subset of the same source dataset (Open-I or MIMIC) as the clients that is not a part of client data. For the server dataset with domain gap, we utilize a different CXR dataset (See Appendix $D$). 
For each of these settings, we first leverage \textbf{FedDF} \cite{lin2020ensemble} that uses ensemble distillation to train a single student model via guidance from multiple teacher models where each teacher represents the updated local model from each client. The distillation is done using KL divergence by constraining the student model to yield the same output logits as the average logits generated by the teacher models. 

Furthermore, we propose a \textbf{Leave-one-out teacher (LOOT)} model to finetune each client model in the server by enforcing constraints in the feature representation space targeted towards matching the embeddings of other client models. To this end, we define a mean cosine similarity matrix across all models in a mini-batch based on the embeddings and finetune a given client model (student model) by maximizing the similarity of its mean embeddings with respect to that of the other client models (teacher models). For given $K$ models coming from local updates in K clients, we leave one model (which is used as student model) and use $K-1$ other models as teacher model to bridge the gap between the models. This is performed for all the $K$ client models. 

\begin{figure}[t]
    \centering
\includegraphics[width=1.02\columnwidth]{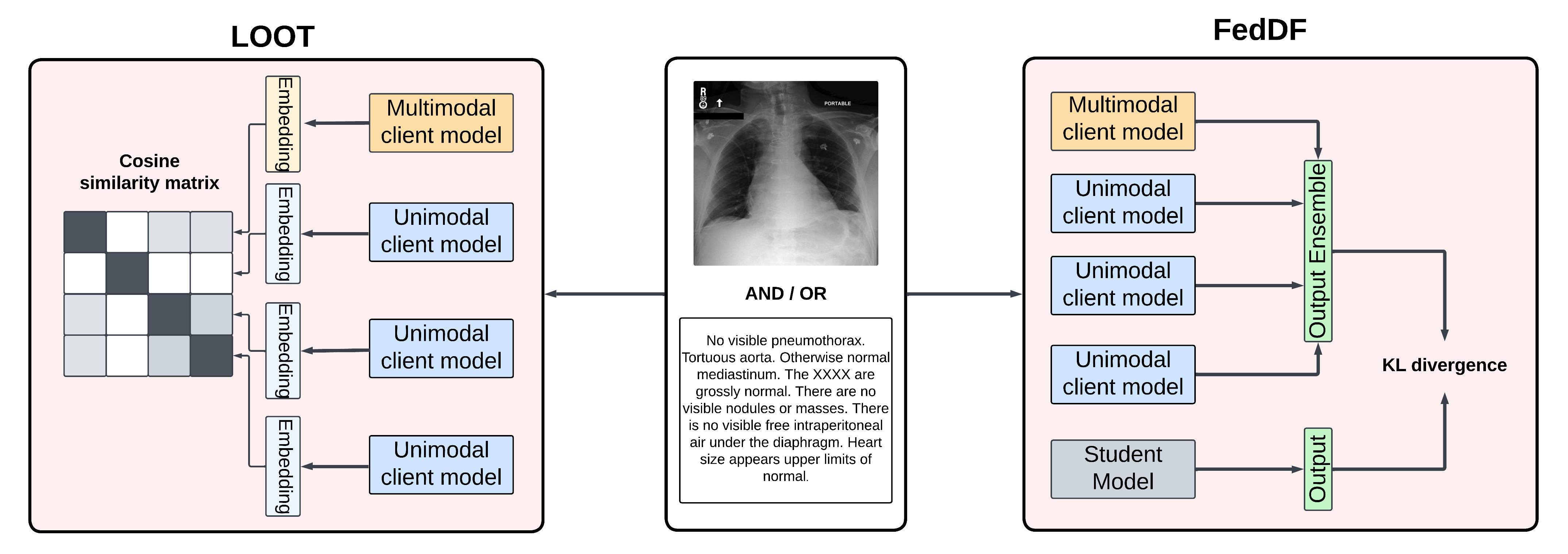}
    \caption{Server-level solutions - LOOT vs FedDF} 
\label{fig:5}
\end{figure}
\subsection{Performance analysis}
As shown in Table 4, FL algorithms like FedProx, MOON, and FedDF perform slightly better than FedAvg in dealing with modality heterogeneity. The proposed multimodal extensions, enforcing a stronger constraint, further improve the accuracy in each case. MAD+ consistently outperforms the other client-based methods as it leverages a locally-expert teacher model pre-trained on its own client to provide additional supervision on the unimodal task. The server-based models utilizing both image and text of the additional unlabeled data in the server performs better than others. LOOT (T+I) consistently performs better than all other methods as it fine-tunes each client model by trying to match the embeddings of other client models. It is particularly effective as it enforces the unimodal clients to produce multimodal-like embeddings that reduces the modality incongruity effects. Another interesting observation is that \color{blue}LOOT is the only model capable of surpassing the unimodal FL performance for both M:U=3:1 and 1:3 for $\gamma=0.5$. However, for $\gamma=0.1$, while LOOT still achieves the best performance, it cannot outperform the corresponding unimodal FL. \color{black}

\section{Conclusion}
This paper investigates the issue of modality incongruity in MMFL in the context of multilabel disease classification from CXR and radiology reports. Instead of proposing complex training procedures, we take the less-explored path of analysing simple yet effective and practical ways of solving the problem. Our investigation is based on better utilization of existing techniques from FL literature, adaptation of known methods from other areas as well as introduction of novel yet intuitive MMFL methods. Our comprehensive evaluation demonstrates that modality imputation is the most effective and practical method in terms of tackling modality heterogeneity, closely followed by server-level finetuning of the client models leveraging unlabeled data on server (See Appendix D and E for more details).

\section{Impact Statement}

Our work on multi-modal federated learning presents a transformative approach to medical imaging, particularly in the context of chest radiography, carrying profound implications for diverse healthcare settings. It addresses the pressing challenges of a diminishing number of radiologists in urban areas and the scarcity of radiology services in resource-poor regions, highlighting the potential to revolutionize healthcare outcomes globally. By leveraging federated learning, the algorithm enables collaborative model training across institutions without compromising data privacy, thereby addressing delays and backlogs in medical imaging interpretation. The exploration of scenarios with varying modalities, including missing ones, underscores a commitment to real-world data variability and the mitigation of modality incongruity. This approach not only enhances model generalization but also fosters collaborative learning irrespective of resource differences, adapting to challenges faced by resource-poor environments. 

Central to the research is an exploration of practical healthcare scenarios where clients possess diverse data modalities, ranging from uni-modal to multi-modal data. The key inquiry revolves around whether uni-modal clients, with only one specific modality, should engage in federated learning independently or collaboratively train with counterparts possessing all modalities. This critical examination addresses the pragmatic challenges encountered in real-world healthcare systems, where institutions vary in imaging capabilities. By exploring the efficacy of federated learning when uni-modal and multi-modal clients collaborate or train within their respective groups, the research seeks to unravel optimal approaches for model training and knowledge exchange. This consideration is pivotal for tailoring machine learning strategies to the inherent variability of healthcare institutions, fostering adaptability, and maximizing the collective potential of diverse datasets. The findings contribute to refining the implementation of federated learning in healthcare, holding promise for enhancing diagnostic accuracy and efficiency across institutions with varying modalities. 

By integrating both imaging and textual data (radiology reports), this approach can potentially lead to more accurate and comprehensive disease diagnoses. This is particularly significant in medical settings where accurate diagnosis is critical for effective treatment planning. The use of multimodal data can help in identifying subtle or complex conditions that might be missed when only one type of data is used. The inclusion of two publicly available datasets, MIMIC-CXR and NIH-Open-I, suggests a commitment to developing solutions that are accessible and applicable to a wide range of healthcare settings, including those with limited resources. This can help in reducing disparities in healthcare access and quality, especially in under-resourced or rural areas where advanced diagnostic tools might not be readily available. The development of such advanced diagnostic tools can also contribute to medical research by providing more detailed and comprehensive data for study. Furthermore, it can be used as an educational tool for medical students and professionals, enhancing their understanding of disease presentations and improving their diagnostic skills.

While the use of real-world data sets is beneficial for the development of robust models, it also raises concerns regarding patient privacy and data security. Ensuring that patient data is used ethically and securely is paramount. This work adheres to strict data protection regulations and ethical guidelines. There is a risk that healthcare providers may become overly reliant on automated diagnostic tools, potentially leading to a decline in traditional diagnostic skills. Hence, it is important to use such technology as a supplement to, rather than a replacement for, professional medical judgment.

Nevertheless, the development of such technologies could lead to cost savings in the healthcare sector by reducing the time and resources needed for diagnosis. However, it also requires initial investments in technology and training, which could be a barrier for some healthcare providers. 

Addressing a long-tailed multi-label disease classification problem indicates that the technology is designed to recognize both common and rare diseases. This is particularly important for the diagnosis of rare diseases, which often go undetected or misdiagnosed. In conclusion, this work has the potential to significantly impact various aspects of society, particularly in enhancing healthcare quality and accessibility. However, it is crucial to balance the benefits with ethical considerations and the potential risks associated with the use of advanced technology in healthcare.

\bibliography{icml2024/example_paper}
\bibliographystyle{icml2024}

\newpage
\appendix
\onecolumn

\section{Dataset Details}

We use two different datasets in this study,\textit{ viz.}, MIMIC-CXR and Open-I. 
MIMIC-CXR is a comprehensive dataset comprising 227,835 imaging studies conducted on 65,379 patients who visited the Beth Israel Deaconess Medical Center Emergency Department from 2011 to 2016. The dataset includes a total of 377,110 images, with most studies typically containing both frontal and lateral views. Only frontal views have been utilized in this work. Additionally, the dataset provides semi-structured free-text radiology reports written by practicing radiologists at the time of routine clinical care.

The Open-I dataset, also known as Indiana University (IU) X-ray dataset, contains 7,466 images out of which 3,851 are paired with diagnostic radiology reports. We selected a total of 3,547 frontal view image-report pairs from this dataset.

The class distribution of the datasets is shown in Fig. 6 and 7. As evident from the figures, the class distribution of the datasets is widely different from each other. MIMIC CXR only shows mild imbalance whereas Open-I shows severe imbalance. Fig. 8 shows 24 sample Chest X-Ray images from the MIMIC CXR dataset. As evident from the figure, the dataset exhibits significant variability in terms of image quality, positioning, and patient characteristics. This variability makes it challenging to develop a robust and generalizable model that can handle diverse imaging conditions. Besides, the available disease labels are slightly noisy as they are extracted based on a natural language processing tool called Chexpert labeler from the text radiology reports. Fig. 9 shows the sample reports of MIMIC CXR that consist of a number of sections each - examination, indication, comparison, findings, and impression. Only the "Findings" section of the report were used in this study.

24 sample Chest X-Ray images from the Open-I dataset have been shown in Fig. 10. As evident, the images are remarkably different from that of the MIMIC dataset. While MIMIC CXR is primarily derived from a clinical database of intensive care unit (ICU) patients, Open-I includes images from different clinical contexts, not necessarily limited to ICU patients. However, the number of samples in the dataset is limited which makes it harder to train deeper models on this dataset without overfitting. Table 5 shows the findings section of 17 randomly chosen sample reports along with their labels. As observed from the table, the length of the reports can vary widely depending on the patient case and radiologist and can correspond to one or more disease categories.

\begin{figure}
    \centering
\includegraphics[width=1.05\columnwidth]{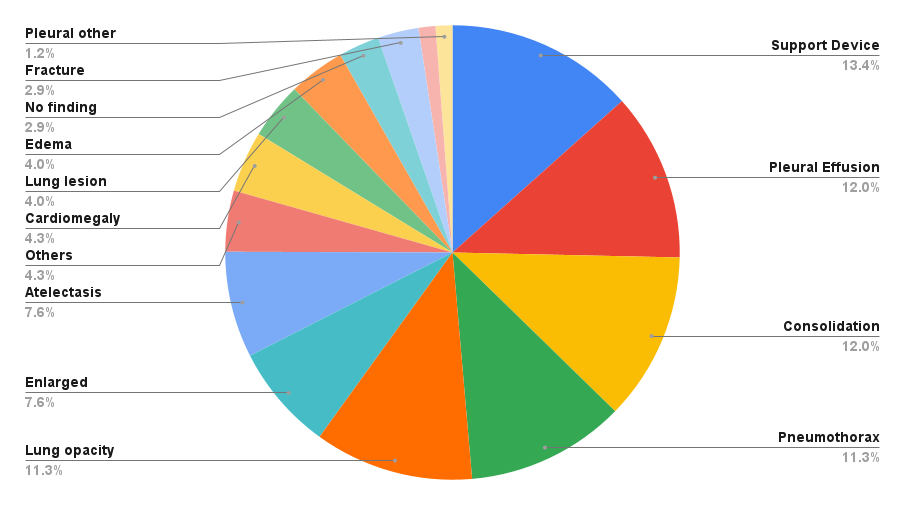}
    \caption{Class proportions (in terms of percentage) in MIMIC Chest X-Ray dataset} 
\label{fig:7}
\end{figure}

\begin{figure}
    \centering
\includegraphics[width=1.05\columnwidth]{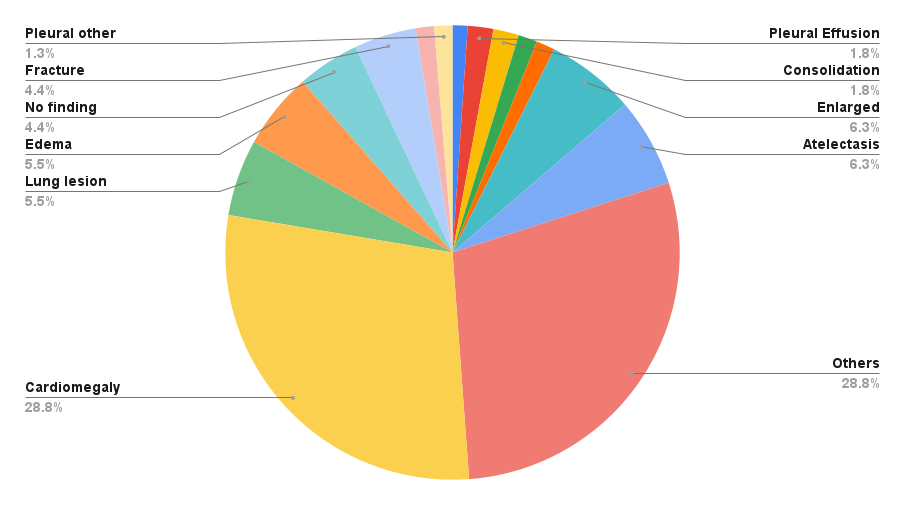}
    \caption{Class proportions (in terms of percentage) in Open-I Chest X-Ray dataset} 
\label{fig:8}
\end{figure}

\begin{figure}
    \centering
\includegraphics[width=0.9\columnwidth]{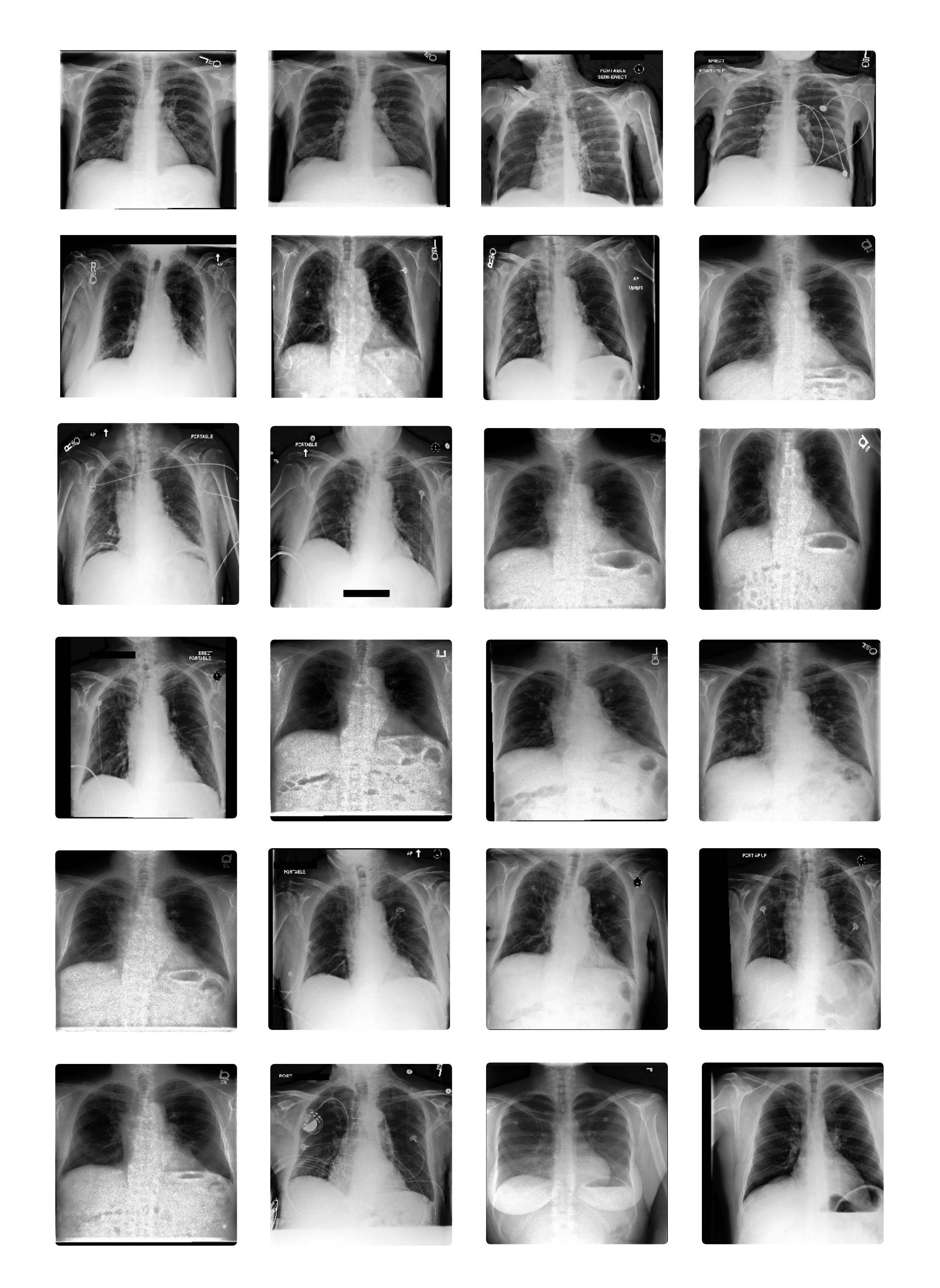}
    \caption{Sample Chest X-Ray images from MIMIC CXR dataset} 
\label{fig:9}
\end{figure}

\begin{figure}
    \centering
\includegraphics[width=0.9\columnwidth]{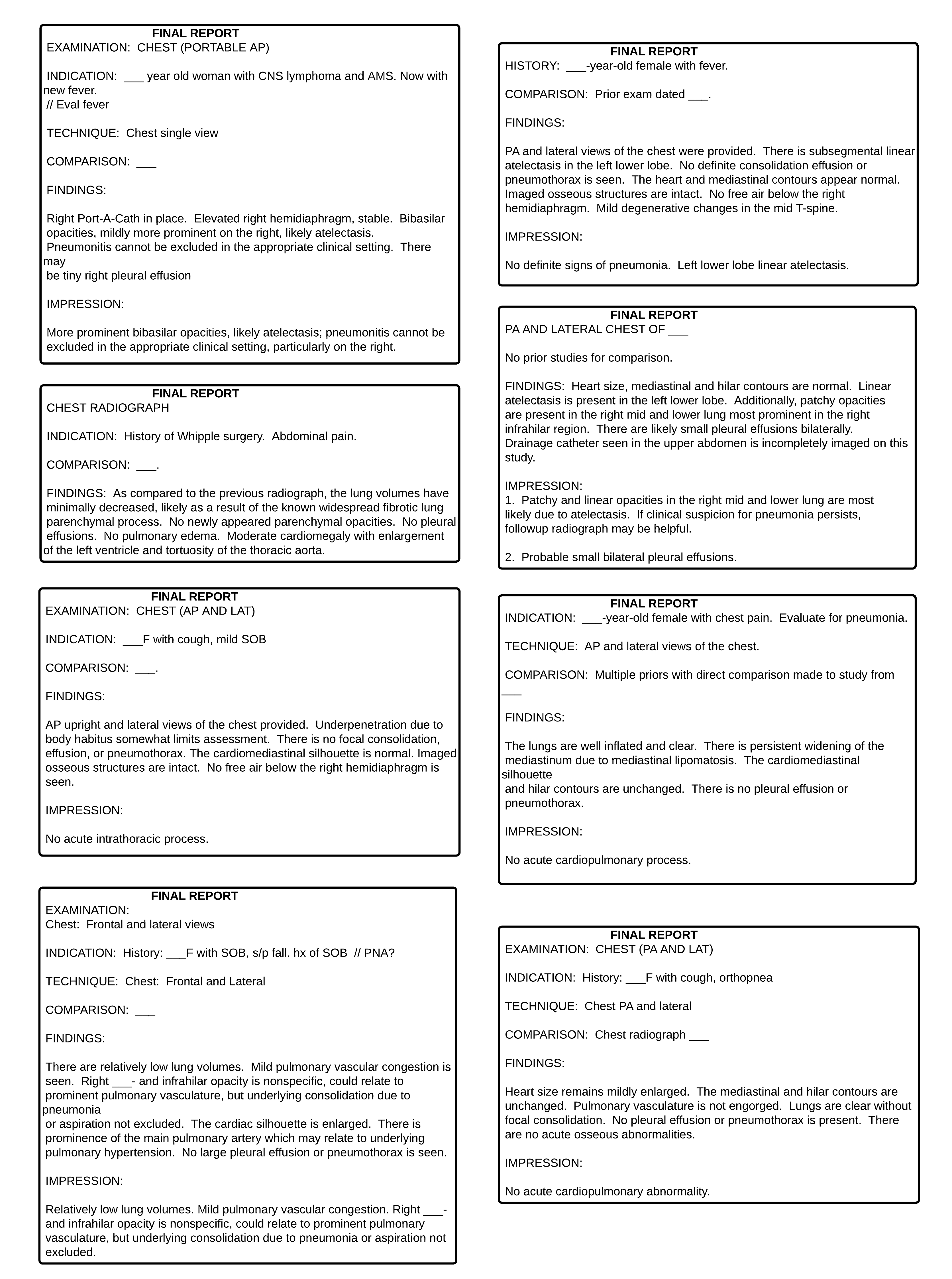}
    \caption{Sample reports from (Multimodal) \textbf{CLIENT 1}. In this work, we only use the "FINDINGS" section as radiology report (text modality).} 
\label{fig:10}
\end{figure}

\begin{figure}
    \centering
\includegraphics[width=0.9\columnwidth]{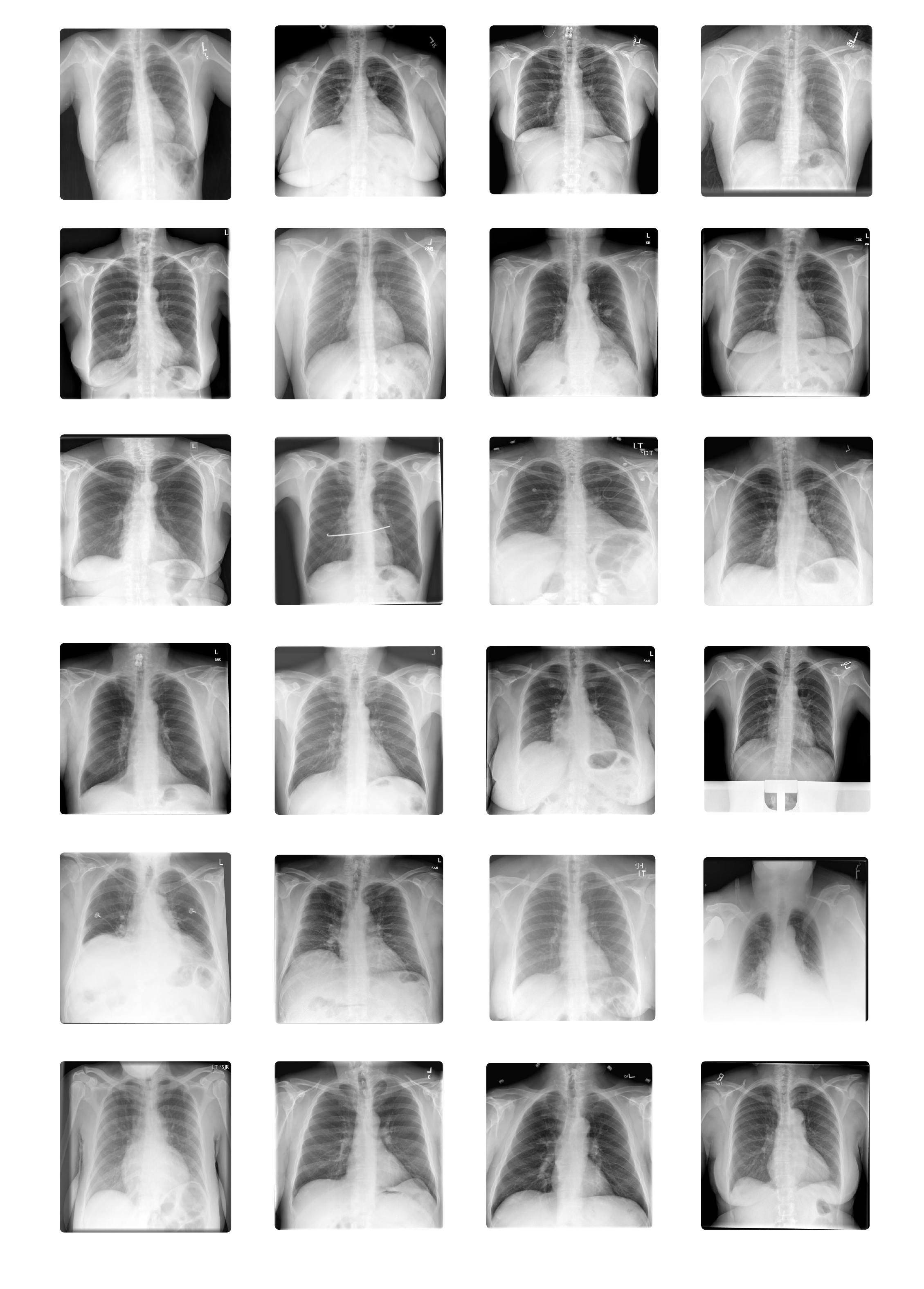}
    \caption{Sample Chest X-Ray images from Open-I dataset. As evident, the images are diverse and contain several artifacts. The images are also notably different from MIMIC-CXR.} 
\label{fig:11}
\end{figure}

\begin{table*}[htbp]
    \centering
    \caption{The findings of sample reports from Open-I dataset along with the corresponding labels} 
    \scalebox{0.7}{
    \begin{tabular}{|l|l|}
    \hline    
        \textbf{Reports} & \textbf{Labels} \\ \hline
Stable appearance of the right aortic XXXX. Normal heart size. No pneumothorax, pleural effusion or suspicious focal airspace opacity. & No Finding \\ \hline
The heart, pulmonary XXXX and mediastinum are within normal limits. There is no pleural effusion or pneumothorax. There is a region & 
\\of left upper lobe perihilar opacity identified. & Lung Opacity\\ \hline
The cardiomediastinal silhouette and pulmonary vasculature are within normal limits in size and contour. There is a XXXX-A-XXXX &  \\terminating at the caval atrial junction, without evidence of pneumothorax. There is no focal airspace disease. There are small calcified &\\nodules in the superior segment of the right lower lobe, XXXX old granulomatous infection. There are no acute bony findings  &Pneumonia\\\hline
No focal consolidation, pneumothorax, or pleural effusions. Stable calcified granulomas. Cardiomediastinal silhouette demonstrates mild  & Enlarged \\tortuosity of the thoracic aorta and atherosclerotic calcifications of the aortic XXXX. No acute osseous abnormality identified. &Cardiomediastinum\\\hline
In the interval, consolidation and atelectasis have developed in the right lower lobe. Costophrenic XXXX blunted on the right. & Consolidation,  \\
Left lung clear. Heart size normal.  &Atelectasis \\\hline

Chest  Comparison: There is a 2.6 cm diameter masslike density over the lingula partial obscuration left cardiac XXXX. There may be some &\\ ill-defined opacity in the right mid and lower lung zone. No pleural effusion is seen. The heart is borderline enlarged. The aorta is dilated and &\\tortuous. Arthritic changes of the spine are present.  Pelvis and left hip There is an impacted and rotated fracture through the neck of the femur &\\ on the left. No pelvic fracture is seen. Arthritic changes are present in the lower lumbar spine. Large amount of stool and XXXX obscures &\\ portions of the pelvis.  Femur The femoral images do not XXXX the area of the hip fracture. The remaining portions of the femur appear to be &\\intact with no fracture or destructive process. Extensive atherosclerotic vascular disease throughout the superficial femoral artery is present.  &\\Left knee There is osteoporosis and mild arthritic changes. No fracture is seen. No dislocation is identified. Severe atherosclerotic changes &Cardiomegaly,\\of the superficial femoral and popliteal artery are seen. &  Lung Lesion \\\hline
No heart size is normal. The lungs are clear. No nodules or masses. Bilateral nipple shadows seen overlying the anterior 6th ribs. Minimal fibrosis &\\in the right apex, may be due to XXXX radiation treatment.& Pleural other \\\hline
Stable postoperative changes with midline sternotomy XXXX and myocardial revascularization. Cardiac size remains mildly enlarged but stable.&Cardiomegaly,\\There is mild vascular congestion. Small bilateral pleural effusions are present, which are XXXX. &  Pleural Effusion \\\hline

Prominent hiatal hernia as before. Anticipated senescent changes of mediastinum. Opacity seen XXXX on lateral XXXX XXXX involving both &Enlarged\\right middle lobe and lingula compatible with some bronchiectasis and chronic inflammatory change. There may be some chronic indolent & Cardiomediastinum, \\ infection here associated with some chronic consolidation. Perhaps some slight progression, but overall XXXX change since prior examination. &Lung Opacity,\\On lateral view, the posterior lung bases are grossly clear. No effusions or CHF. &  Consolidation\\\hline

The lungs are hyperinflated with mildly coarsened interstitial markings consistent with chronic lung disease. No focal consolidation, pneumothorax, &\\or effusion identified. The mediastinal silhouette is stable and within normal limits for size. There is redemonstration without significant change in&\\ right hilar calcified lymph XXXX. The bony structures of the thorax demonstrate degenerative changes of the right shoulder and a XXXX right&\\ humerus consistent with distal humeral amputation. No acute bony abnormality identified. & Lung Opacity\\\hline

No comparison chest x-XXXX XXXX lungs. Lucency left chest compatible with relatively large pneumothorax and collapse of substantial &\\portion of left lung. No substantial mediastinal shift seen. Right lung grossly clear.& Pneumothorax\\ \hline

Stable right-sided subclavian central venous catheter with tip approximating the SVC. Stable right suprahilar opacity, compatible with history of &\\right upper lobe mass. Elevation of the right hemidiaphragm. Right-sided pneumothorax noted measuring approximately 1.8 cm from the the right &Lung Opacity,\\apex. Stable postsurgical changes left axilla. Degenerative changes thoracic spine. Stable streaky opacities right base. XXXX opacity right midlung,& Pneumothorax,\\ question fluid level, incompletely evaluated, no recent XXXX for comparison. & Support Devices \\ \hline

There is stable, mild enlargement of the cardiac silhouette. Stable mediastinal silhouette. There are low lung volumes with bronchovascular &Cardiomegaly,\\crowding. Scattered XXXX opacities in the right lung base XXXX representing foci of subsegmental atelectasis with scattered airspace opacities &Lung Opacity,\\in the medial left lower lobe. No pleural effusion. Degenerative changes of the thoracic spine possibly consistent with DISH. &  Atelectasis, \\ & Pneumothorax  \\\hline

There is a minimally displaced fracture of the right lateral 7th rib. There is a small right pleural effusion with associated atelectasis of the right lower &\\lobe. There appears to be a healing fracture of the posterolateral right 8th rib. There is questionable cortical defect involving the sternum seen XXXX &\\on lateral view. XXXX would be XXXX to evaluate this finding. As the small right-sided pleural effusion is visible on both PA and lateral views. &Atelectasis,\\There is a XXXX left-sided pleural effusion as well. The left lung appears grossly clear. Heart size and pulmonary XXXX appear normal. &Pleural Effusion,\\There is a mild scoliosis involving the thoracic spine.   &Fracture\\ \hline

On the right there is marked narrowing of the hip joint space uniformly throughout. Osteophyte formation is present with some sclerosis and&\\ subchondral cyst formation vertically along the superior acetabulum and femoral head. I do not see evidence for fracture or destructive process. &\\AP view of the femur shows no femoral XXXX destructive process or other significant abnormality. For of the Left hip shows near-complete &\\obliteration of the joint space with severe subchondral sclerosis and cystic formation in both the superior acetabulum and superior aspect of the femoral &\\head. No fracture or destructive process is identified. Surgical markers were XXXX in the images and left hip for the purpose of surgical planning. PA &\\and lateral chest show the lungs to be clear. There may be some hyperinflation. No pleural effusion is identified. The heart is normal in size. There are&\\ calcified mediastinal lymph XXXX. The skeletal structures appear normal.  &Support Devices\\ \hline
Chest: 2 images. Heart size is normal. Mediastinal contours are maintained. There is a mild pectus excavatum deformity. The lungs are clear of focal &\\infiltrate. There is no evidence for pleural effusion or pneumothorax. No convincing acute bony findings. Right shoulder: 3 images. There has been&\\ XXXX and screw fixation of the midshaft right clavicle. The lateral most screw is fractured. This is age-indeterminate as no prior studies are available&Enlarged \\ for comparison. Otherwise, the surgical XXXX appears intact. The humeral head is seen within the glenoid, without evidence for dislocation. &Cardiomediastinum,\\No bony fractures are seen. The visualized right ribs appear intact. Right clavicle: 2 images. No clavicle fracture is seen. Once again noted is the &Fracture,\\surgical fixation XXXX, with fracture of the lateral most fixation screw.  &Support Devices\\ \hline

Chronic bilateral emphysematous changes. The heart size and mediastinal silhouette are within normal limits for contour. The lungs are clear. No &\\pneumothorax or pleural effusions. The XXXX are intact. Stable splenic artery embolism coils. &Support Devices\\ \hline

The lungs are clear. Heart and pulmonary XXXX appear normal. The pleural spaces are clear and mediastinal contours are normal. Nodular density& Lung lesion,\\ overlying the anterior left 4th rib XXXX represents a healing rib fracture. & Fracture\\ \hline

    \end{tabular}}
\end{table*}

\section{Implementation Details}
\textbf{Our code will be made publicly available upon the acceptance of the paper.} We employ a pre-trained ResNet-50 model, originally trained on the ImageNet dataset, as our visual feature extractor. The input images are resized to a dimension of $256 \times 256$ pixels with three color channels. We extract visual features from the last feature map of ResNet-50, which has dimensions of $16\times16\times2048$. During pre-training, we randomly sample 180 visual features, resulting in a feature matrix of size $180\times2048$. However, for FL-based classification task, we utilize all available features, resulting in a matrix of size $256 \times 2048$. 

To incorporate text tokens into our model, we process report sequences by truncating or padding them to a fixed length of 253 tokens, taking into account the maximum embedding size. For the joint embedding of visual and textual information, we adopt a BERT-base architecture consisting of 12 Transformer layers. Each Transformer layer comprises 12 attention heads, a hidden size of 768 for embedded representations, and a dropout probability of 0.1. We utilize the AdamW optimizer with a learning rate set to 1e-5 for both the visual backbone and the Transformer. Training is performed on NVIDIA RTX A6000 GPU (48 GB) with a batch size of 32, and we conduct local training for 50 epochs to pre-train the model for avoiding overfitting. For the pre-training of the transformer and the alignment of visual features with language features, we employ two tasks: Masked Language Modeling (MLM) and Image Report Matching (IRM) in the multimodal clients. We set the number of workers as 4, width of vision embeddings as 768, width of text embeddings as 768, total embedding size as 768, hidden embedding size as 768, gradient accumulation steps as 4, vocabulary size as 30522, and a weight decay of 0.01. The total number of rounds for FL is kept as 100 with 1 local epoch in each round. We use binary cross entropy with logits loss as the loss function for the multilabel classification task.

We use $\gamma=0.1$ and $\gamma=0.5$ to simulate heterogeneous data partition for FL clients. For server-level methods, we use held-out $5\%$ of the total data samples available. These are used as unlabeled samples. For FedProx, FedMultiProx, MOON, and MultiMOON, we tuned the regularization coefficient via grid search from 0.001 to 0.1 and fixed the optimal value as $\lambda=0.01$. The temperature ($\tau$) was varied from 0.1 to 1.0 at an interval of 0.1 and was fixed at 0.5.

For modality imputation network, we use a 12-layered BERT-base model with 12-head attention in each layer and train it for 100 epochs. We use a learning rate of 5e-6 with weight decay of 1e-6. We keep the embedding drop-out, feedforward drop-out, and post-attention drop-out as 0.1. We use causal attention mask with local window size = 256.

\section{Formulation of Self-attention Mechanism}
The self attention matrix (SA) is computed using the query and key vectors of the joint embedding and can be denoted as:
\begin{equation}
    SA=\left[\begin{array}{ccccc}
S_q \cdot S_k & \cdots & S_q \cdot W_{1 k} & \cdots & S_q \cdot E_{k} \\
V_{1 q} \cdot S_k & \cdots & V_{1 q} \cdot W_{1 k} & \cdots & V_{1 q} \cdot E_{k} \\
\vdots & \ddots & \vdots & \ddots & \vdots \\
S E P_{q} \cdot S_k & \cdots & S E P_{q} \cdot W_{1 k} & \cdots & S E P_{q} \cdot E_{k} \\
W_{1 q} \cdot S_k & \cdots & W_{1 q} \cdot W_{1 k} & \cdots & W_{1 q} \cdot E_{k} \\
\vdots & \ddots & \vdots & \ddots & \vdots \\
W_{K q} \cdot S_k & \cdots & W_{K q} \cdot W_{L k} & \cdots & W_{K q} \cdot E_{k} \\
E_{q} \cdot S_k & \cdots & E_{q} \cdot W_{L k} & \cdots & E_{q} \cdot E_{k}
\end{array}\right]
\end{equation}

Therefore, based on modality type, self attention matrix (SA) can be expressed in terms of four subparts: 
\begin{equation}
S A_{q, k}  =S A_{S_q: S E P_{q}, S_k: S E P_{k}} 
+S A_{S_q: S E P_{q}, W_{1 k}: E_{L k}} 
+S A_{W_{1 q}: S E P_{q}, S_k: S E P_{k} }
+S A_{W_{1 q}: E_{q}, W_{1 k}: E_{k}}
\end{equation}

Accordingly, the isolated attention mask is denoted as:
\begin{equation}
    Iso_M=\left[\begin{array}{cccccc}
0 & \cdots & 0 & -\infty & \cdots & -\infty \\
\vdots & \ddots & \vdots & \vdots & \ddots & \vdots \\
0 & \cdots & 0 & -\infty & \cdots & -\infty \\
-\infty & \cdots & -\infty & 0 & \cdots & 0 \\
\vdots & \ddots & \vdots& \vdots & \ddots & \vdots \\
-\infty & \cdots & -\infty & 0 &  \cdots & 0 \\
\end{array}\right] \in \mathbb{R}^{L_{emb} \times L_{emb}}
\end{equation}

The causal attention mask is denoted as:
\begin{equation}
    Causal_M=\left[\begin{array}{cccccc}
0 & \cdots & 0 & -\infty & \cdots & -\infty \\
\vdots & \ddots & \vdots & \vdots & \ddots & \vdots \\
0 & \cdots & 0 & -\infty & \cdots & -\infty \\
0 & \cdots & 0 & 0 & \cdots & -\infty \\
\vdots & \ddots & \vdots& \vdots & \ddots & \vdots \\
0 & \cdots & 0 & 0 &  \cdots & 0 \\
\end{array}\right] \in \mathbb{R}^{L_{emb} \times L_{emb}}
\end{equation}

The Partial Bidirectional attention mask is denoted as:
\begin{equation}
    parBi_M=\left[\begin{array}{cccccc}
0 & \cdots & 0 & 0 & \cdots & 0 \\
\vdots & \ddots & \vdots & \vdots & \ddots & \vdots \\
0 & \cdots & 0 & 0 & \cdots & 0 \\
0 & \cdots & 0 & 0 & \cdots & -\infty \\
\vdots & \ddots & \vdots& \vdots & \ddots & \vdots \\
0 & \cdots & 0 & 0 &  \cdots & 0 \\
\end{array}\right] \in \mathbb{R}^{L_{emb} \times L_{emb}}
\end{equation}

The Bidirectional attention mask can be denoted as:
\begin{equation}
    Bi_M=\left[\begin{array}{cccccc}
0 & \cdots & 0 & 0 & \cdots & 0 \\
\vdots & \ddots & \vdots & \vdots & \ddots & \vdots \\
0 & \cdots & 0 & 0 & \cdots & 0 \\
0 & \cdots & 0 & 0 & \cdots & 0 \\
\vdots & \ddots & \vdots& \vdots & \ddots & \vdots \\
0 & \cdots & 0 & 0 &  \cdots & 0 \\
\end{array}\right] \in \mathbb{R}^{L_{emb} \times L_{emb}}
\end{equation}

\section{Additional Experimental Results}
\subsection{Report generation and downstream performance of MIN}
Figures 11, 12, and 13 show the qualitative performance of the modality imputation network in generating radiology reports in each of the 3 unimodal clients - C2, C3, and C4 respectively. The corresponding BLEU-4 scores have been provided in the Table 3 of the main paper. In each of these figures, we demonstrate five or six randomly chosen samples with their original report and generated report placed next to each other for the ease of comparison. The figures show that MIN is very effective in imputing the missing text modality. There is large overlap between the generated and original images. A careful study of these figures reveals that sometimes, the language of generated text closely resembles that of the Ground Truth report, while at other times the expressions are different but eventually refer to the same disease or phenomenon. Eg: While the ground truth report in the first example of Fig. 11 says "the heart size is within normal limits", generated report says "cardiomediastinal silhouette is normal". Similarly, the original report states "there are no acute osseous abnormalities" whereas generated report says "imaged osseous structures are intact." More such examples can be found in the Figures 11-13. The overlapping areas have been marked using red font for the ease of understanding.

These reports are then used as "proxy" reports to fill the gap of missing modality. The quality of the generated reports is further demonstrated through the downstream task of 14 class disease classification. Figure 29 shows the per-class accuracy curves (in $\%$) for each of the 14 categories of CXR disease for MIMIC CXR. Each curve denotes a binary classification accuracy curve in the context of multilabel disease classification. The curves show that the overall performance is stable and that convergence is achieved for each category.

\subsection{Visualization of attention maps}

For deeper understanding of multimodal information fusion, we visualize the learned attention maps corresponding to each head of the multiheaded attention blocks separately for each of the 12 layers in case of bidirectional self-attention in Figures 14-25. 
In the first two layers (Fig 15 and 16), the attention maps exhibit a relatively uniform or diffuse pattern, where each token in the input sequence attends to a wide range of other tokens with roughly equal weights. In terms of key (horizontal axis), only a part of the text embeddings is seen to be active.
These layers' attentions are typically focused on capturing local dependencies and identifying basic patterns in the input data. It's akin to a "raw" or initial understanding of the relationships between image and text tokens. We observe less structured attention patterns, with some attention heads potentially attending to specific token pairs based on their initial representations. In the third, fourth, fifth, and sixth layers, the attention maps begin to show more structured and refined patterns compared to the first two layers. The attention heads start to specialize in capturing specific types of relationships or dependencies between the text and image. The attention patterns have become more localized, with tokens attending to nearby tokens that are more relevant to their roles in the sequence.

As we progress to deeper the layers, attention maps tend to become more specialized and abstract (see Figs. 23, 24, 25). 
The attention patterns become increasingly fine-grained and specific to the classification task. Some heads are observed to pay attention to specific tokens with critical information, while others suppress or filter out irrelevant tokens.
The visualization of the attention maps clearly show that the model's ability to capture complex relationships and features in the input data improves with each layer. 
\subsection{Further evidence regarding mitigating Modality incongruity}

The impact of modality incongruity can also be assessed by evaluating the distance between multimodal and unimodal client models. We assess the performance of the methods in question by determining the average L2 distance between the unimodal and multimodal models. Essentially, this involves calculating the mean of six sets of mutual L2 distances among the client models for a four client FL setting. The mean distance curves of different methods with varying number of rounds for both the datasets have been presented in Figs. 26, 27, and 28.

As expected, Figure 26 shows that the mean distances corresponding to the models for fully unimodal and fully multimodal settings in case of MIMIC CXR, indicated by dotted lines, are the least and are quite close to each other. The baseline mean model distance is higher than MIN and LOOT under all settings. This shows that the proposed methods are successful in mitigating the modality heterogeneity by bridging the gap between unimodal and multimodal models. Similar performance is noticed for Open-I dataset corresponding to different client ratios and level of heterogeneity as observed in Figs. 27 and 28. For example, in Fig. 27, LOOT and MIN are seen to decrease the mean model distance from 0.5 to 0.23. Similarly in Fig. 28, the mean model distances decrease from 0.4 - 0.5 to 0.2 - 0.3. All these evidences show MIN and LOOT to be effective in mitigating modality incongruity as claimed in the conclusion of the main paper.

\subsection{Server-level solutions on different dataset and comparison with SOTA}

Due to space constraints in the main paper, we report the results of additional experiments in Table 6 (along with the main results provided in Table 5 for better comparison). Table 6 shows the performance of different server-level models with samples from the same dataset or different dataset on the server. It is to be noted that we only need unlabeled data samples at this step. For the 'different dataset' scenario, we use MIMIC CXR in the server when the clients possess Open-I data and vice versa. As evident from Figs 7-10, the two datasets have a class distribution shift as well as domain shift and hence, indicate realistic situations. This experiment is conducted to evaluate the robustness of the server-level methods against domain gaps. The overall performance of FedDF (I), FedDF (T), and FedDF (I+T) is observed to decrease by a margin of 3.71, 2.22, and 3.29 respectively in terms of AUC. Similarly the overall performance of the proposed LOOT (I), LOOT (T) and LOOT (I+T) decreases by a margin of 3.28, 2.76, and 3.40 respectively, when the server dataset is replaced. The performance degradation is higher for only-image version than the only-text version. This is expected as the radiology report in both the datasets follow a similar style and format whereas the variability is higher in Chest X-Ray images.
We also add the performance of a SOTA method, CreamFL \cite{yu2023multimodal}, that was particularly proposed for MMFL and compare its performance with the other methods as seen in Table 6.
\begin{table*}[t]
    \centering
    \caption{MMFL Performance with server-level solutions. T and I indicate the presence of Text and Image respectively} 
    \scalebox{0.76}{
    \begin{tabular}{|l|ll|ll|ll|ll|ll|ll|ll|ll|}
    \hline 
     &  \multicolumn{8}{|c|}{ \textbf{$\gamma=0.5$}}&  \multicolumn{8}{|c|}{$\gamma=0.1$}\\ \hline
     
       \textbf{Methods} &  \multicolumn{2}{|c|}{\textbf{Open-I}} & \multicolumn{2}{|c|}{\textbf{MIMIC CXR}}&  \multicolumn{2}{|c|}{\textbf{Open-I}} & \multicolumn{2}{|c|}{\textbf{MIMIC CXR}}  &  \multicolumn{2}{|c|}{\textbf{Open-I}} & \multicolumn{2}{|c|}{\textbf{MIMIC CXR}}&  \multicolumn{2}{|c|}{\textbf{Open-I}} & \multicolumn{2}{|c|}{\textbf{MIMIC CXR}} \\ \hline
        \textbf{Methods} & \textbf{AUC} & \textbf{F1} & \textbf{AUC} & \textbf{F1}& \textbf{AUC} & \textbf{F1}& \textbf{AUC} & \textbf{F1} & \textbf{AUC} & \textbf{F1} & \textbf{AUC} & \textbf{F1}& \textbf{AUC} & \textbf{F1}& \textbf{AUC} & \textbf{F1} \\ \hline\hline

         &  \multicolumn{16}{|c|}{ Additional data from the \textbf{SAME} dataset} \\ \hline
            
       FedDF (I)  & 74.03 & 29.36 & 84.88 & 74.82 & 78.29 &  32.03 & 88.20 & 78.90& 61.28 & 27.84 & 77.18 & 70.58 & 65.46 & 28.74 & 82.06 & 73.85 \\
        FedDF (T)  & 71.49 & 26.00 & 82.09 & 72.49 & 77.80 & 30.65 & 85.81 & 77.01 & 59.49 & 26.77 & 73.65 & 71.24 & 63.04 & 27.99 & 80.43 & 74.89 \\
       FedDF (I+T) & 76.72 & 33.94 & 85.13 & 75.45 & 80.22 & 35.18 & 90.82 & 79.90 & 63.93 & \textbf{29.80} & 80.01 & 74.16 & 68.10 & 30.30 & 86.50 & 76.78 \\

        LOOT (I)  & 75.06 & 34.98 & 86.85 & \textbf{78.19} & 79.94 & 33.74 & 90.33 & 81.28 & 62.49 & 28.02 & 82.22 & 74.54 & 66.02 & 28.85 & 86.08 & 75.47 \\
        LOOT (T)  & 73.84 & 27.43 & 83.10 & 71.36 & 78.55 & 31.67 & 89.39 & 79.88 & 60.36 & 27.88 & 78.10 & 71.93 & 65.38 & 28.33 & 83.14 & 74.30 \\
         LOOT (I+T)  & \textbf{79.36} & \textbf{38.73} & \textbf{89.97}  & 77.85 & \textbf{83.75} & \textbf{40.30} & \textbf{92.47} & \textbf{83.34} & \textbf{65.25} & {29.32} & \textbf{84.29} & \textbf{75.17} & \textbf{70.94} & \textbf{30.65} & \textbf{89.60} & \textbf{77.09} \\
          CreamFL & 77.77 & 34.12 & 86.10 & 77.90 & 81.07 &  35.69 & 89.73 & 80.58& 63.11 & 29.18 & 83.00 & 68.70 & 69.19 & 30.11 & 82.49 & 76.15\\
      \hline
        &  \multicolumn{16}{|c|}{ Additional data from the \textbf{DIFFERENT} dataset} \\ \hline
            
       FedDF (I)  & 71.18 & 27.19 & 80.82 & 73.85 & 74.48 &  31.35 & 84.19 & 76.98& 56.10 & 26.55 & 73.60 & 68.11 & 61.32 & 27.23 & 80.01 & 70.19 \\
        FedDF (T)  & 70.05 & 25.56 & 80.28 & 71.87 & 74.66 & 29.94 & 83.20 & 75.19 & 57.00 & 24.79 & 70.70 & 70.48 & 60.09 & 28.07 & 80.05 & 70.76 \\
       FedDF (I+T) & 74.20 & 32.67 & 81.09 & 73.13 & 76.68 & 34.90 & 88.19 & 75.92 & 59.02 & {27.29} & 76.02 & 73.00 & 65.92 & 28.04 & 83.99 & 76.09 \\

        LOOT (I)  & 71.93 & 33.01 & 82.10 & {74.39} & 77.10 & 31.27 & 88.05 & 80.99 & 57.89 & 26.07 & 80.07 & 73.50 & 63.80 & 26.57 & 81.75 & 73.66 \\
        LOOT (T)  & 70.09 & 26.12 & 81.01 & 69.89 & 75.89 & 30.10 & 86.10 & 78.13 & 57.16 & 23.80 & 76.98 & 70.67 & 63.40 & 28.01 & 79.09 & 73.11 \\
         LOOT (I+T)  & \textbf{76.09} & \textbf{37.10} & \textbf{85.10}  & \textbf{76.12} & \textbf{80.13} & \textbf{40.18} & \textbf{89.68} & \textbf{80.83} & \textbf{61.01} & \textbf{{27.98}} & \textbf{81.83} & \textbf{74.09} & \textbf{67.39} & \textbf{30.01} & \textbf{87.18} & \textbf{76.08} \\
          CreamFL & 74.73 & 36.15 & 81.87 & 75.40 & 76.66 & 32.70 & 88.10 & 75.60 & 57.56 & 25.12 & 78.43 & 70.19 & 64.96 & 28.16 & 83.16 & 71.18 \\
      \hline

    \end{tabular}}
\end{table*}

\begin{figure}
    \centering
\includegraphics[width=1.05\columnwidth]{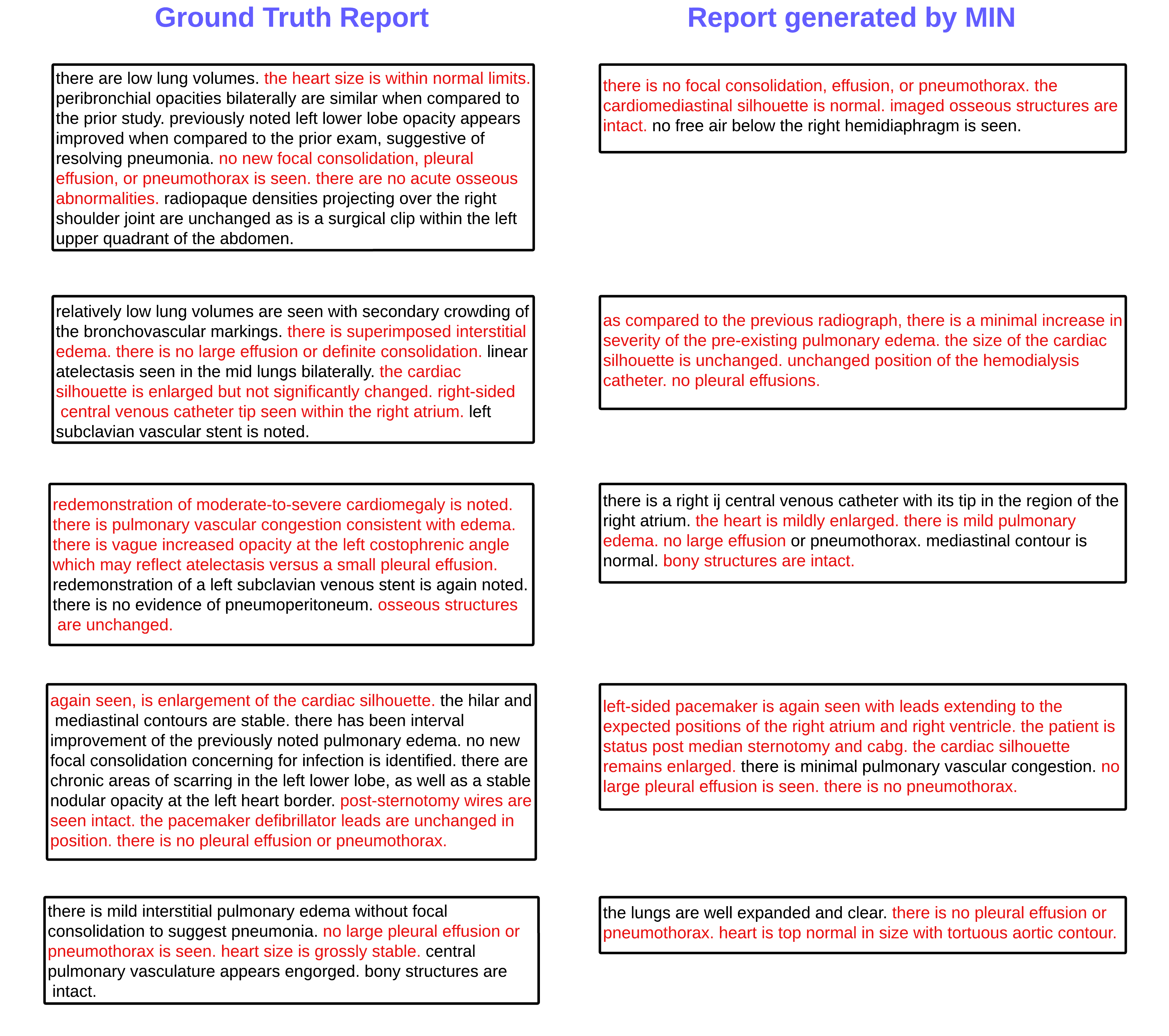}
    \caption{Qualitative performance analysis of Modality Imputation Network (MIN). The figure shows comparison of randomly selected original reports and corresponding generated reports in \textbf{CLIENT 2}. Red font colour indicates similar findings} 
\label{fig:12}
\end{figure}
\begin{figure}
    \centering
\includegraphics[width=1.02\columnwidth]{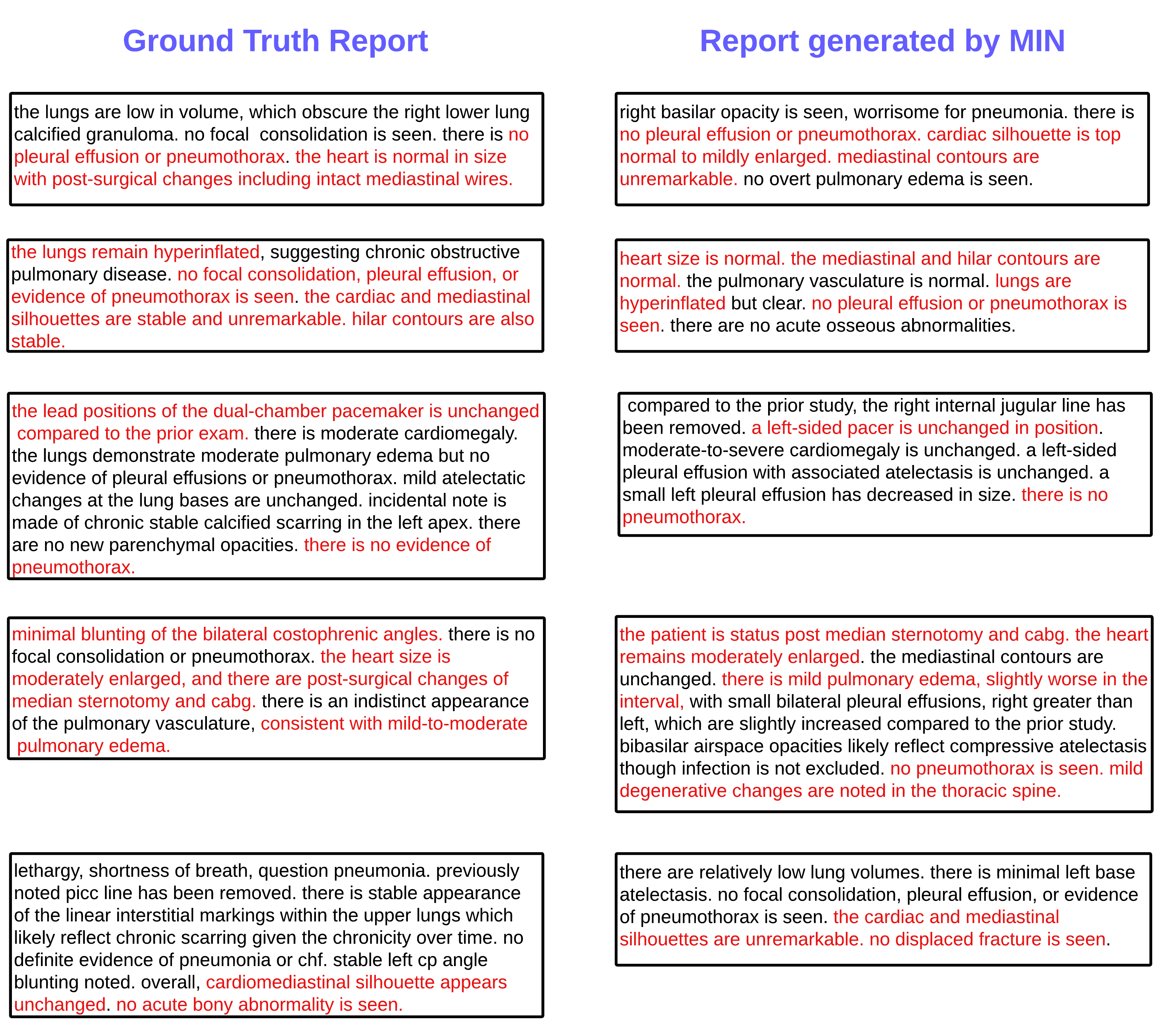}
    \caption{Qualitative performance analysis of Modality Imputation Network (MIN). The figure shows comparison of randomly selected original reports and corresponding generated reports in \textbf{CLIENT 3}. Red font colour indicates similar findings} 
\label{fig:13}
\end{figure}

\begin{figure}
    \centering
\includegraphics[width=1.02\columnwidth]{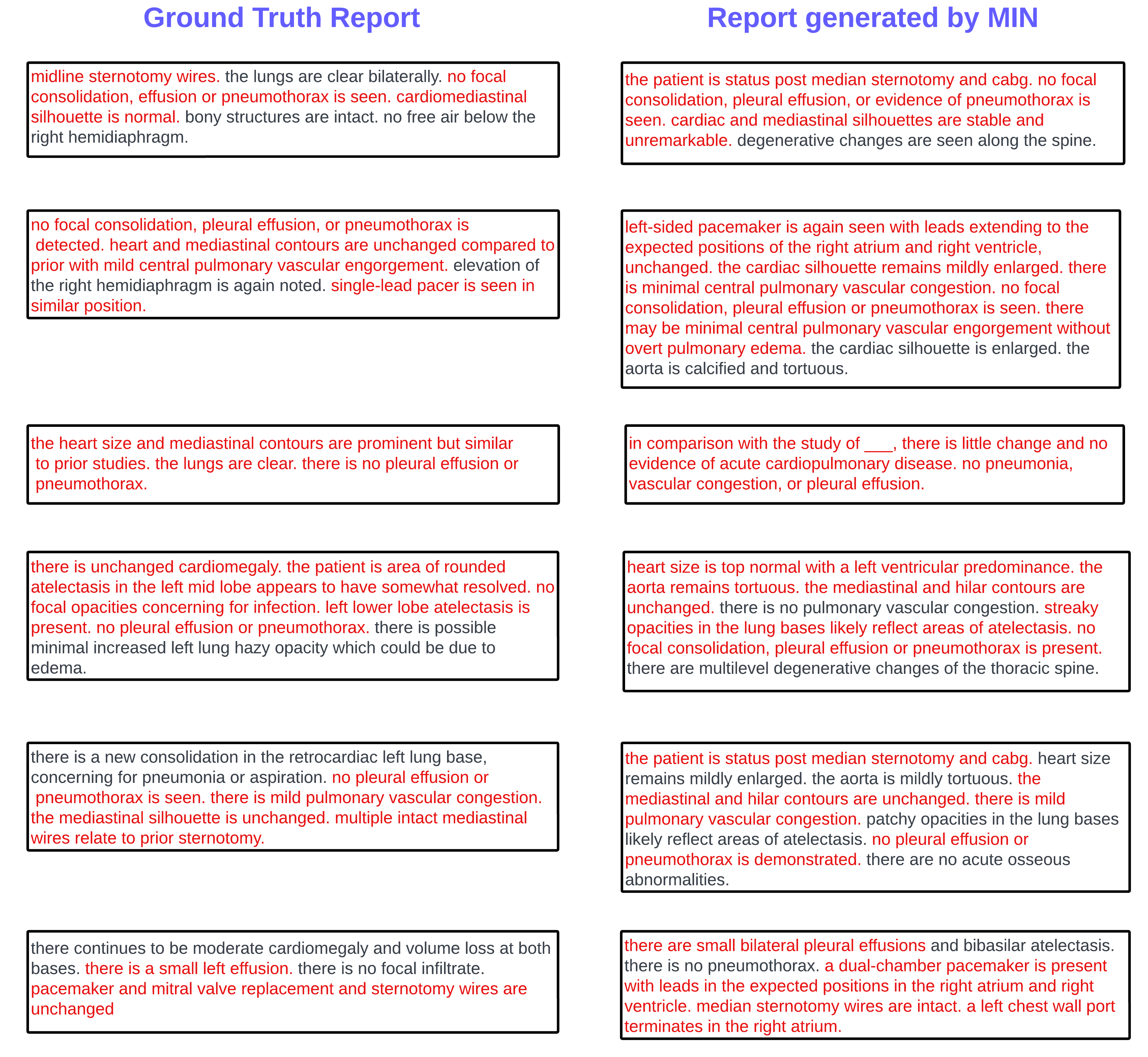}
    \caption{Qualitative performance analysis of Modality Imputation Network (MIN). The figure shows comparison of randomly selected original reports and corresponding generated reports in \textbf{CLIENT 4}. Red font colour indicates similar findings} 
\label{fig:14}
\end{figure}

\begin{figure}
    \centering
\includegraphics[width=1.0\columnwidth]{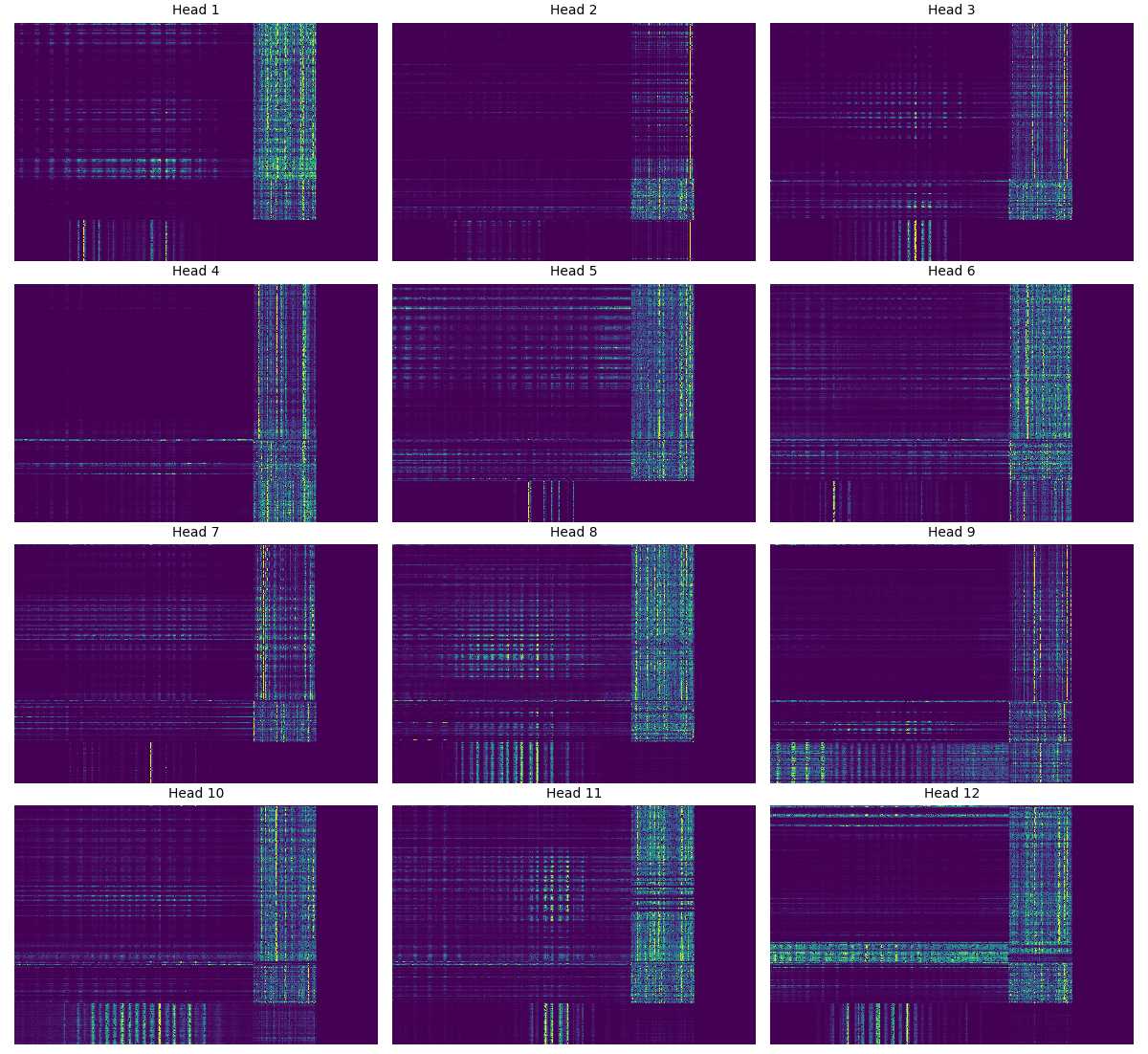}
    \caption{Visualization of Layer 1 Attention Map} 
\end{figure}
\begin{figure}
    \centering
\includegraphics[width=1.0\columnwidth]{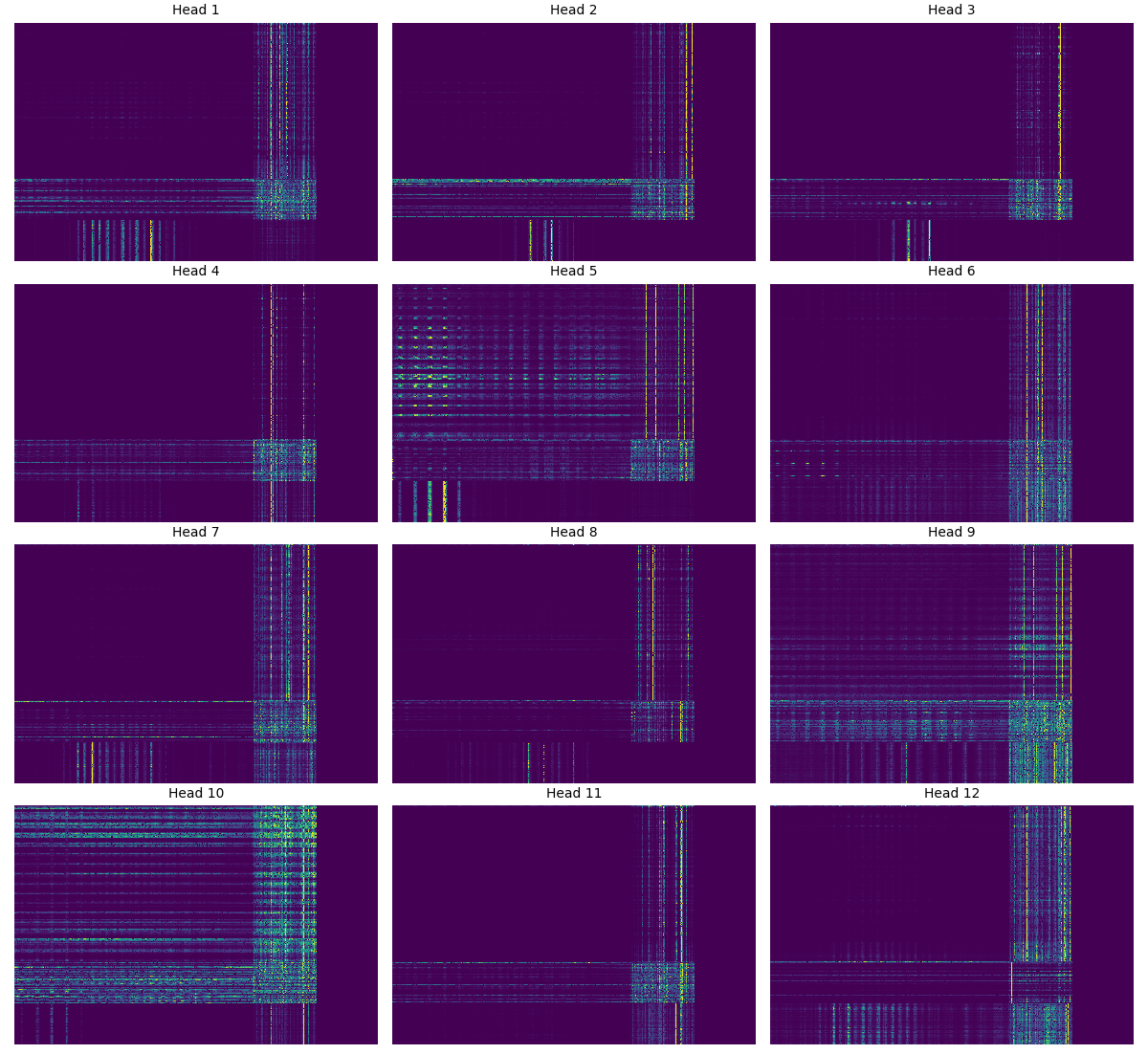}
    \caption{Visualization of Layer 2 Attention Map} 
\end{figure}
\begin{figure}
    \centering
\includegraphics[width=1.0\columnwidth]{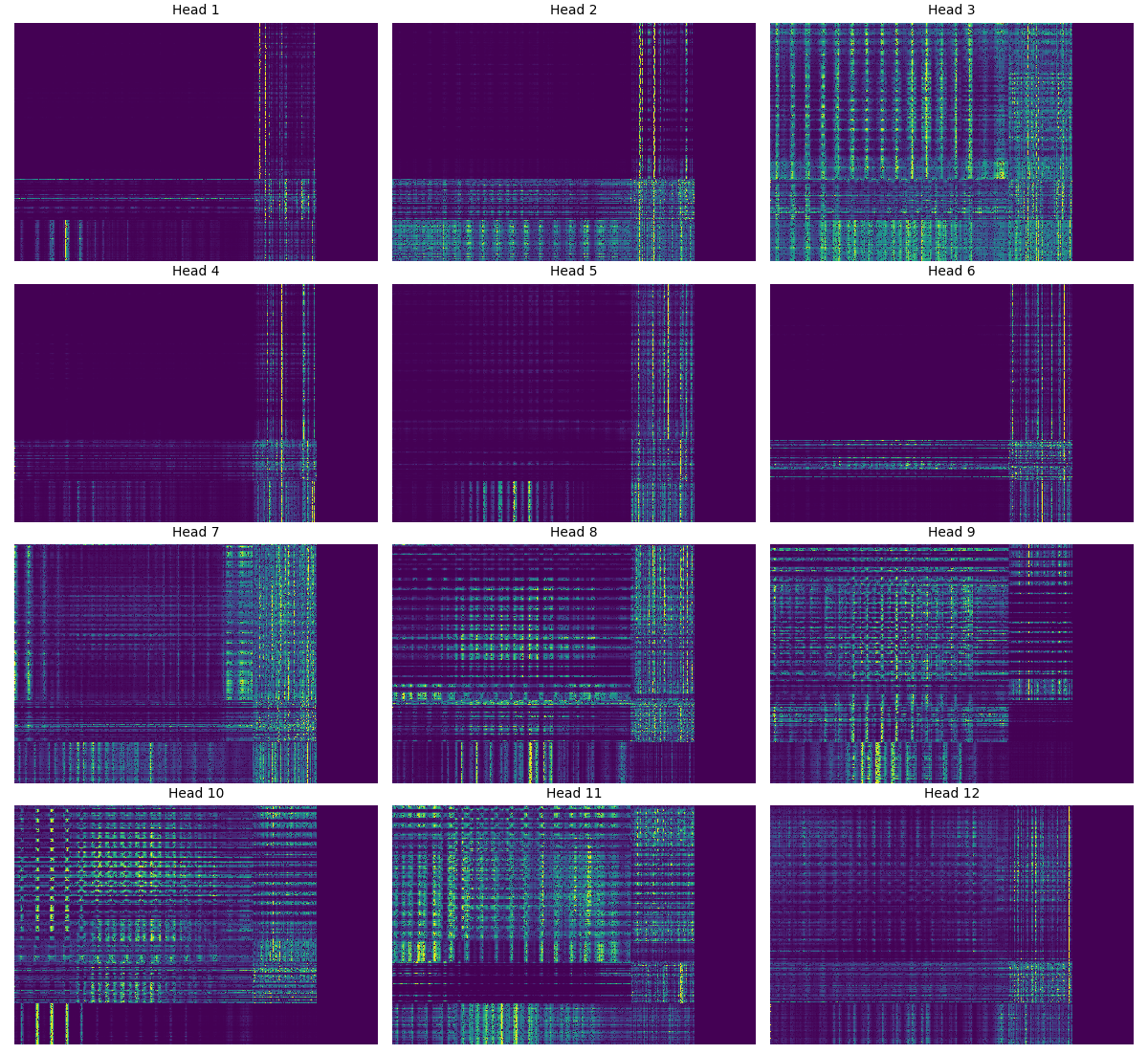}
    \caption{Visualization of Layer 3 Attention Map} 
\end{figure}
\begin{figure}
    \centering
\includegraphics[width=1.0\columnwidth]{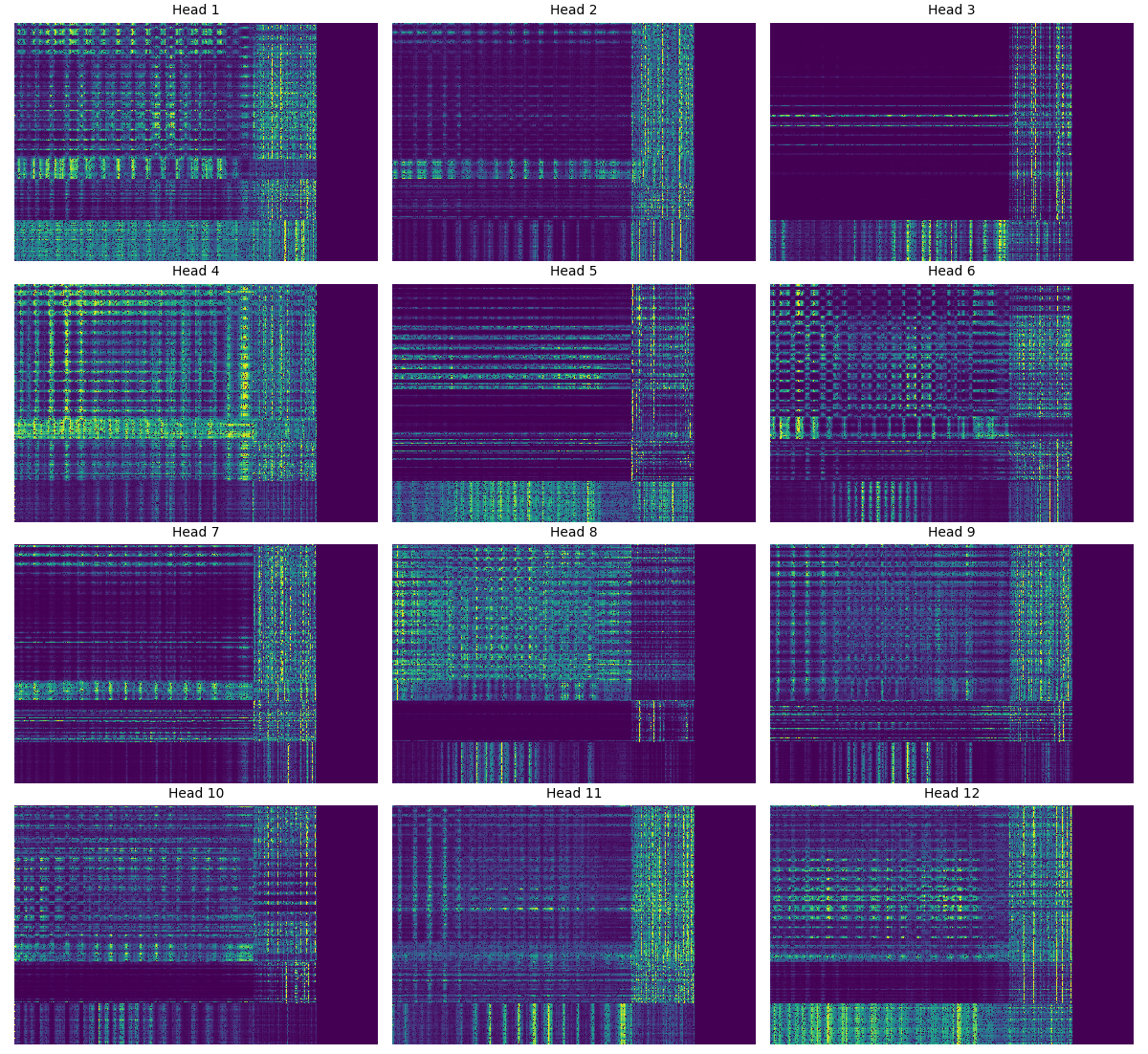}
    \caption{Visualization of Layer 4 Attention Map} 
\end{figure}
\begin{figure}
    \centering
\includegraphics[width=1.0\columnwidth]{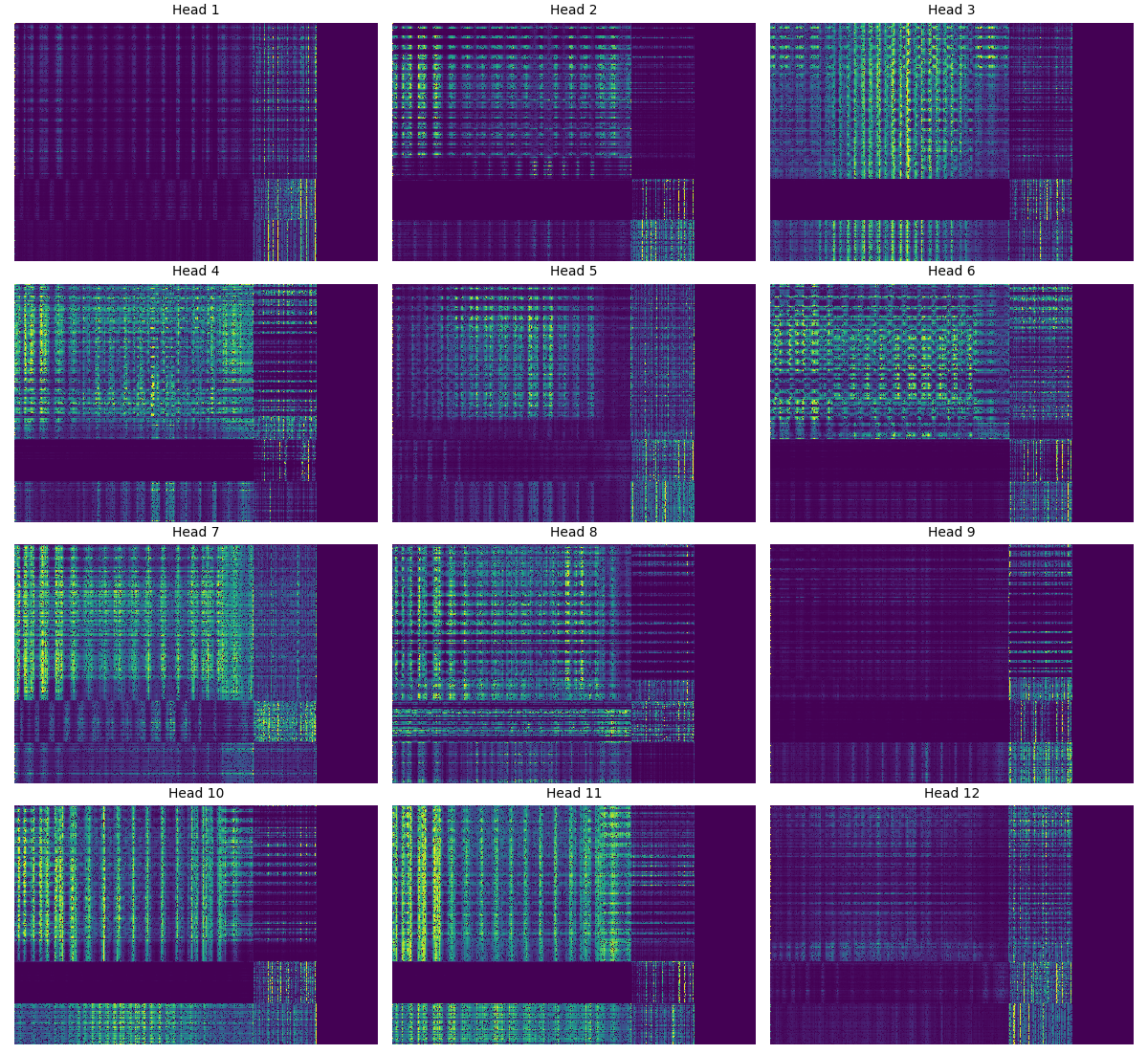}
    \caption{Visualization of Layer 5 Attention Map} 
\end{figure}
\begin{figure}
    \centering
\includegraphics[width=1.0\columnwidth]{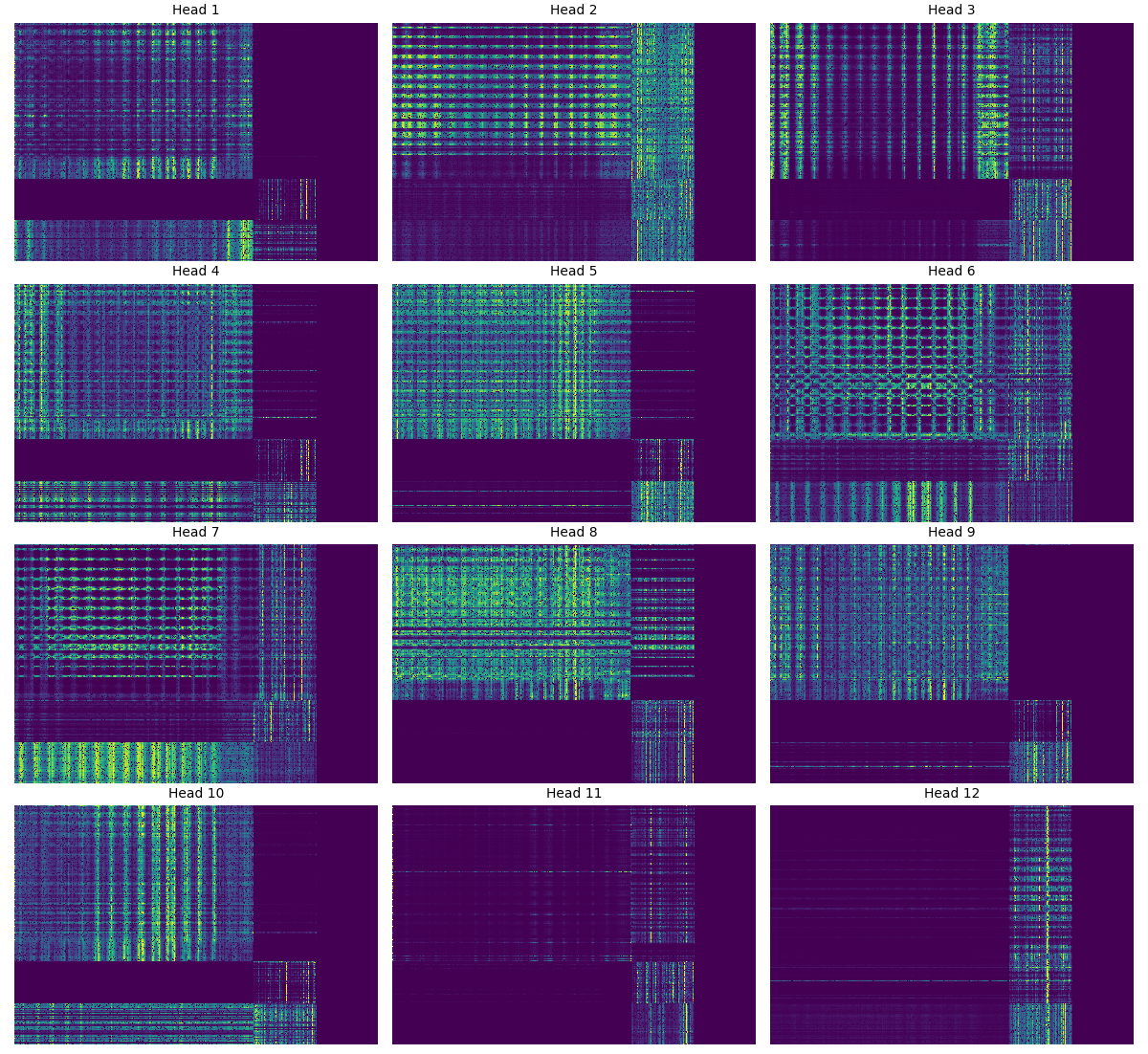}
    \caption{Visualization of Layer 6 Attention Map} 
\end{figure}
\begin{figure}
    \centering
\includegraphics[width=1.0\columnwidth]{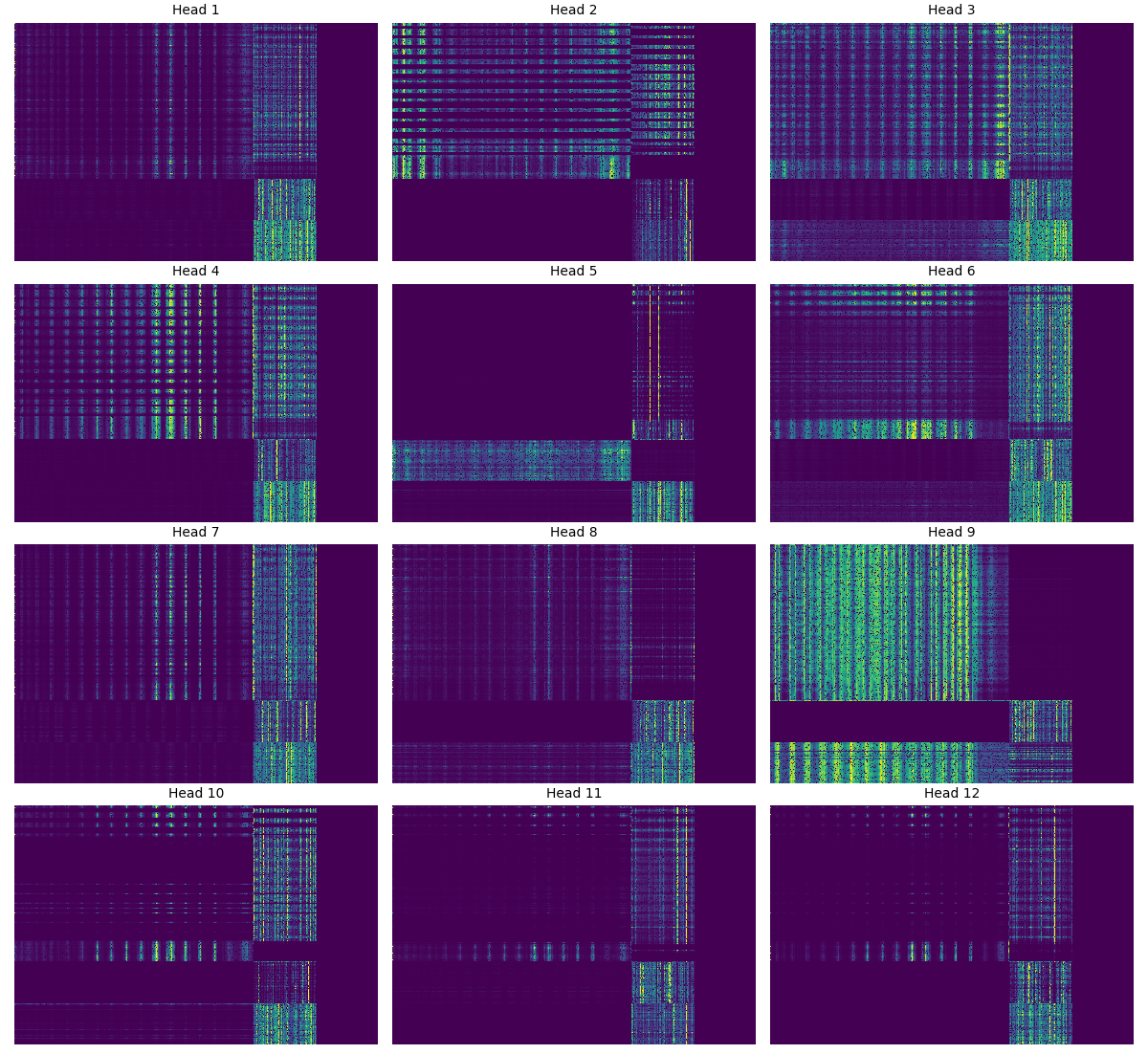}
    \caption{Visualization of Layer 7 Attention Map} 
\end{figure}
\begin{figure}
    \centering
\includegraphics[width=1.0\columnwidth]{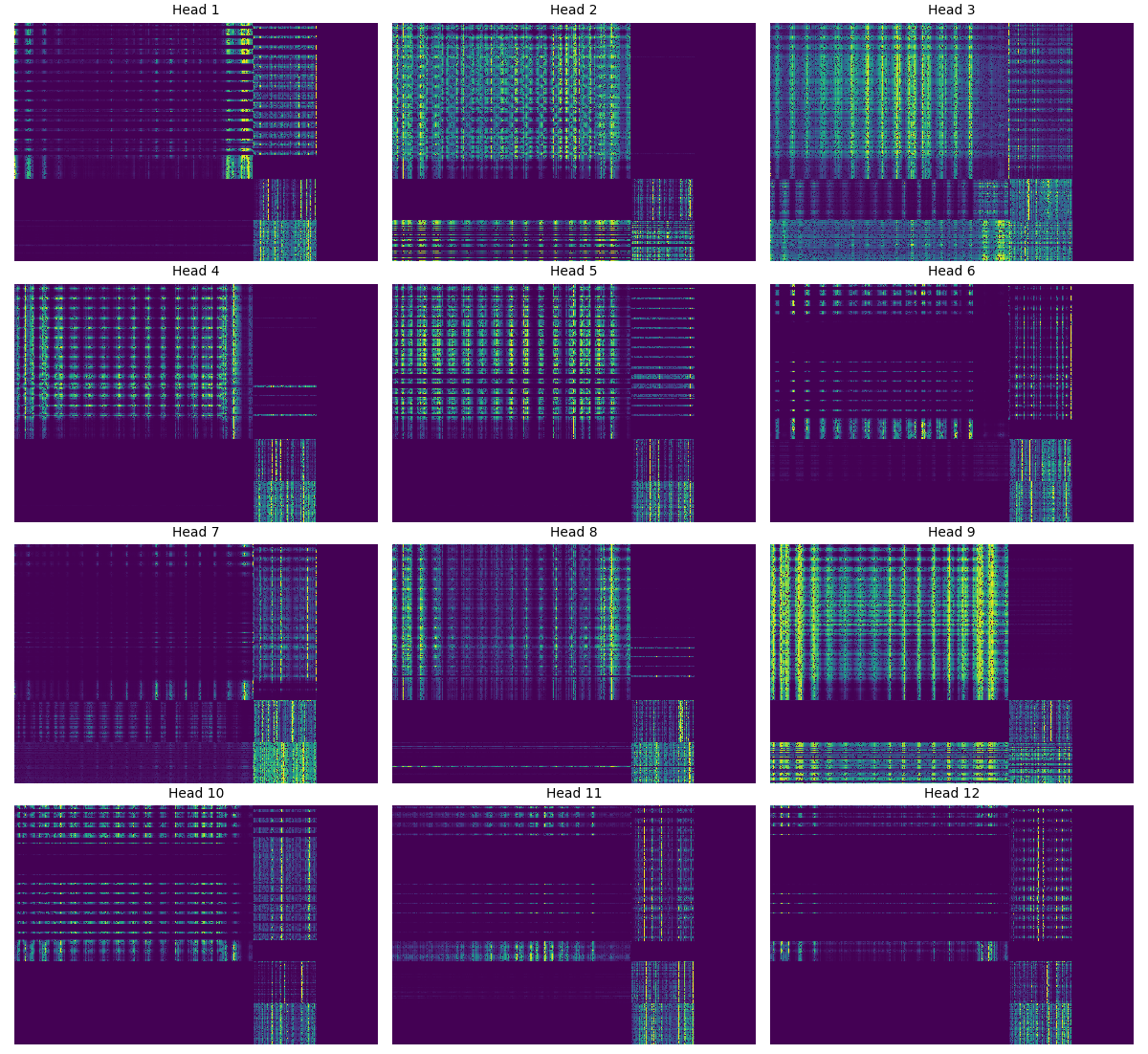}
    \caption{Visualization of Layer 8 Attention Map} 
\end{figure}\begin{figure}
    \centering
\includegraphics[width=1.0\columnwidth]{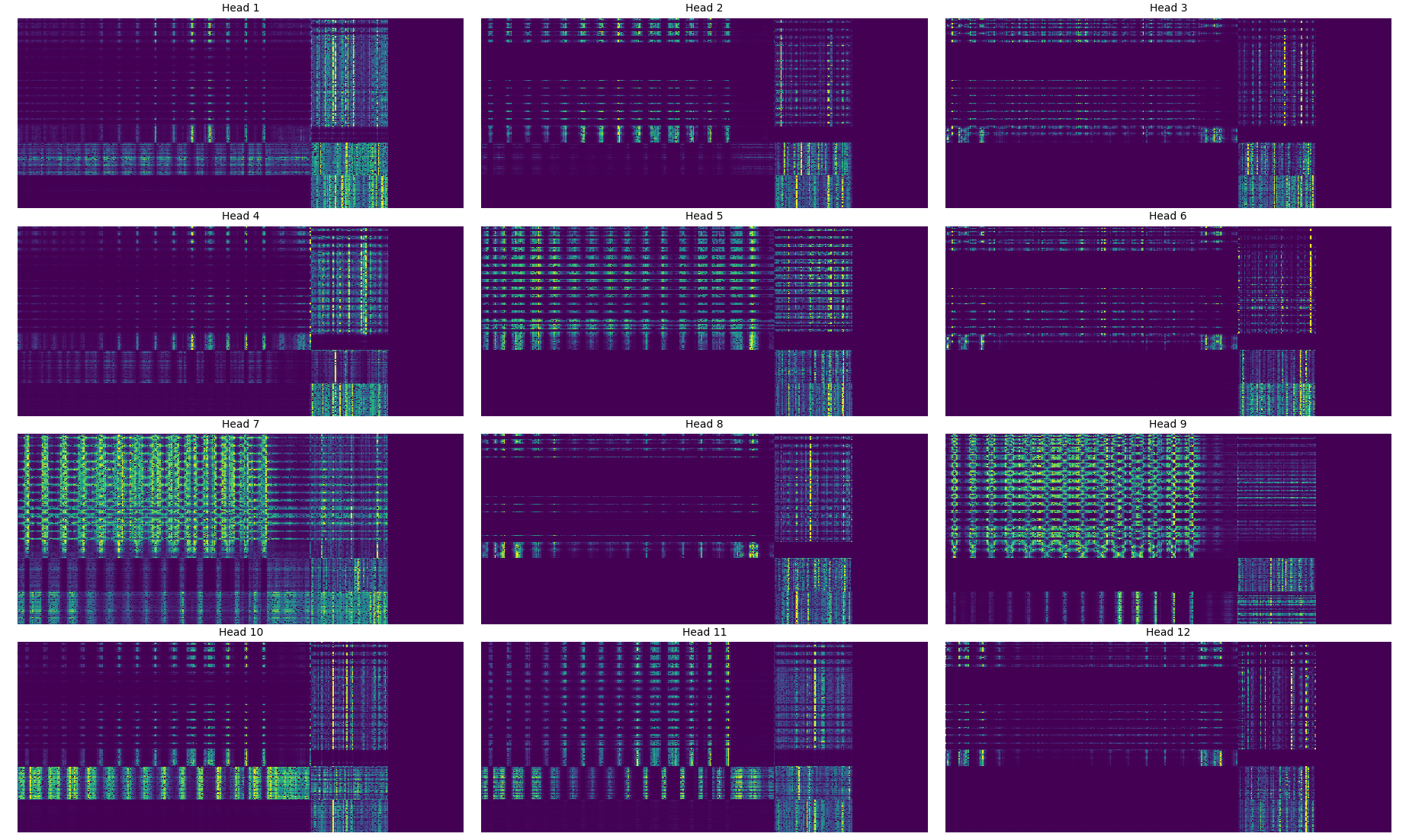}
    \caption{Visualization of Layer 9 Attention Map} 
\end{figure}\begin{figure}
    \centering
\includegraphics[width=1.0\columnwidth]{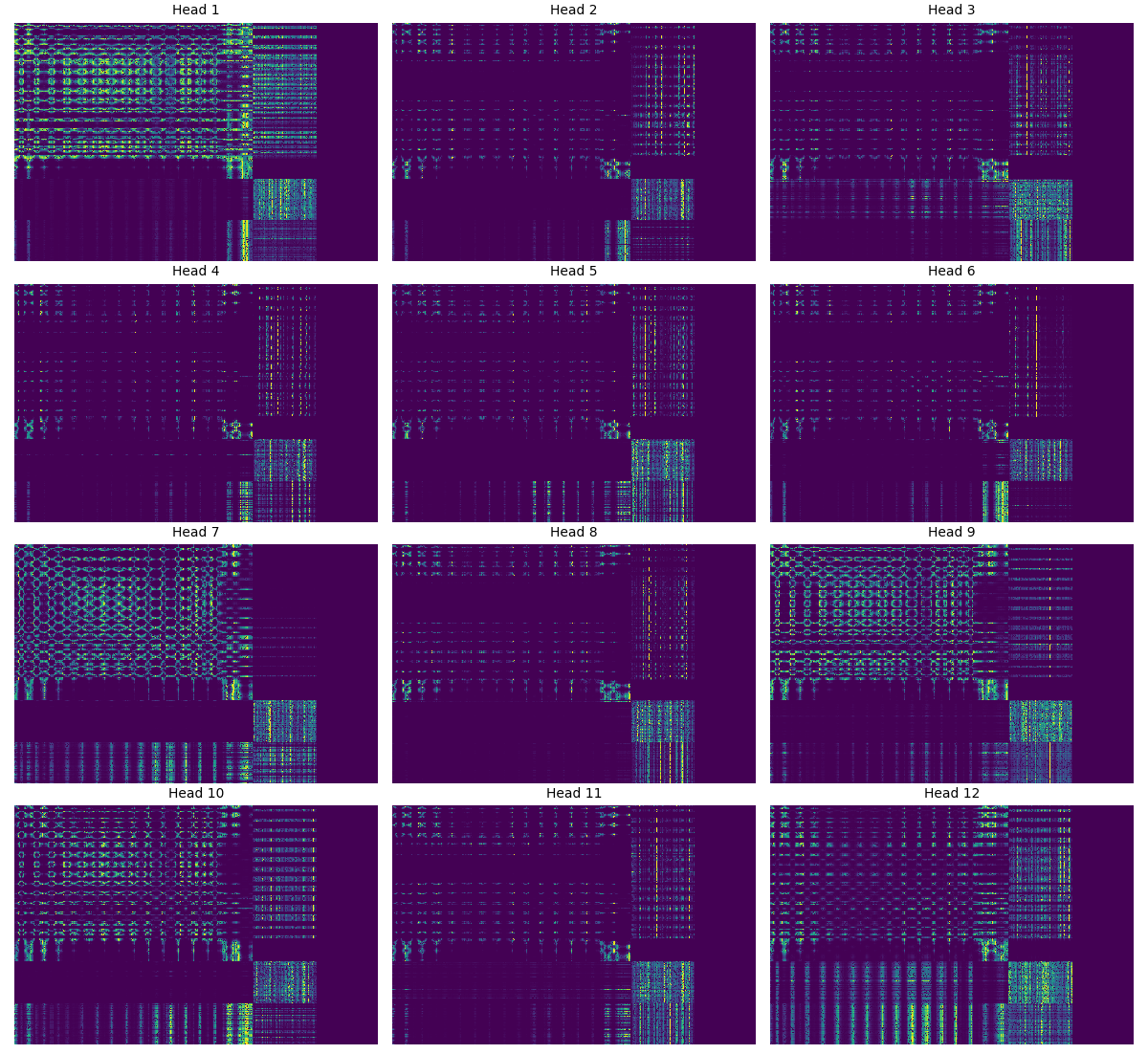}
    \caption{Visualization of Layer 10 Attention Map} 
\end{figure}\begin{figure}
    \centering
\includegraphics[width=1.0\columnwidth]{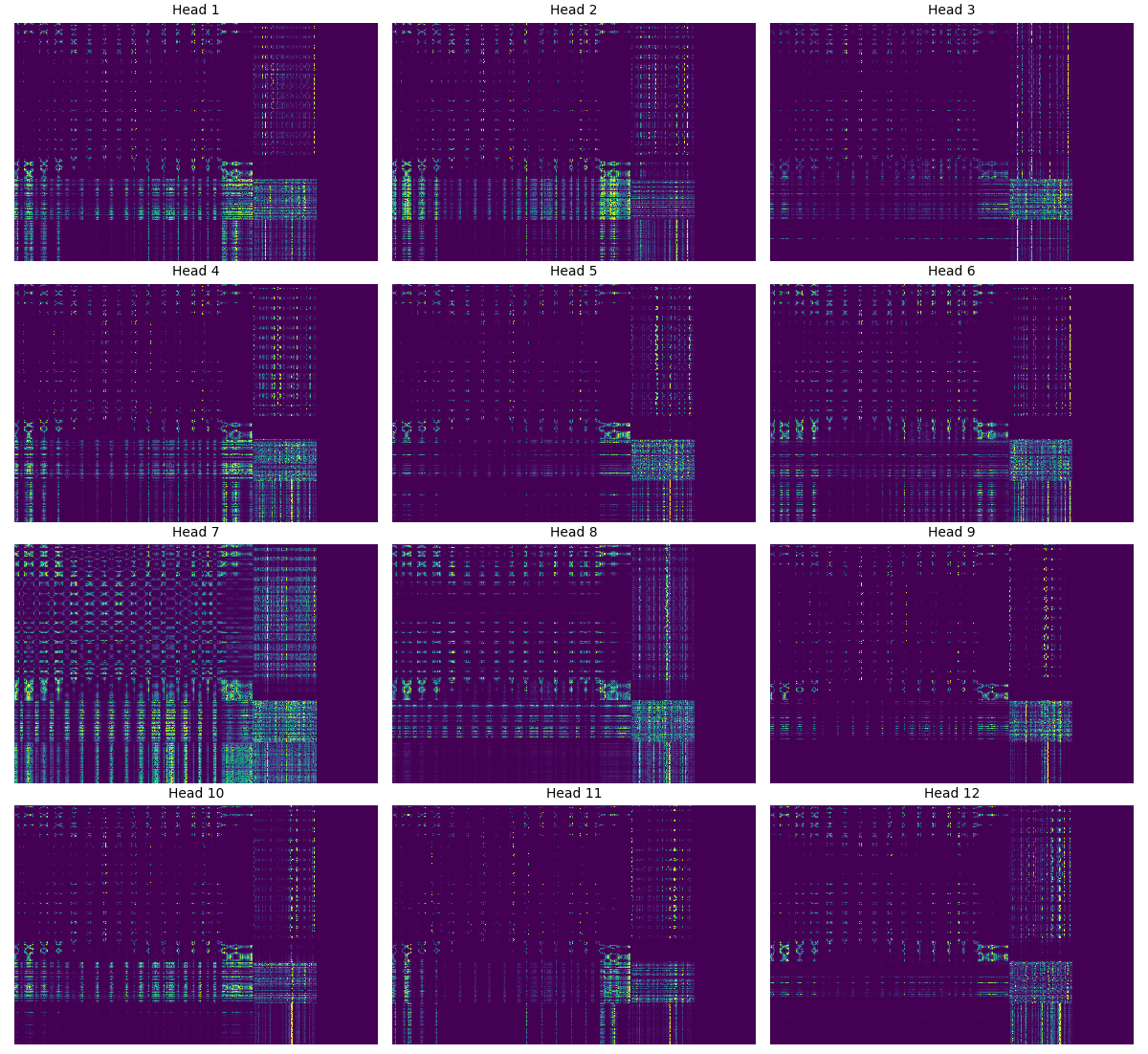}
    \caption{Visualization of Layer 11 Attention Map} 
\end{figure}
\begin{figure}
    \centering
\includegraphics[width=1.0\columnwidth]{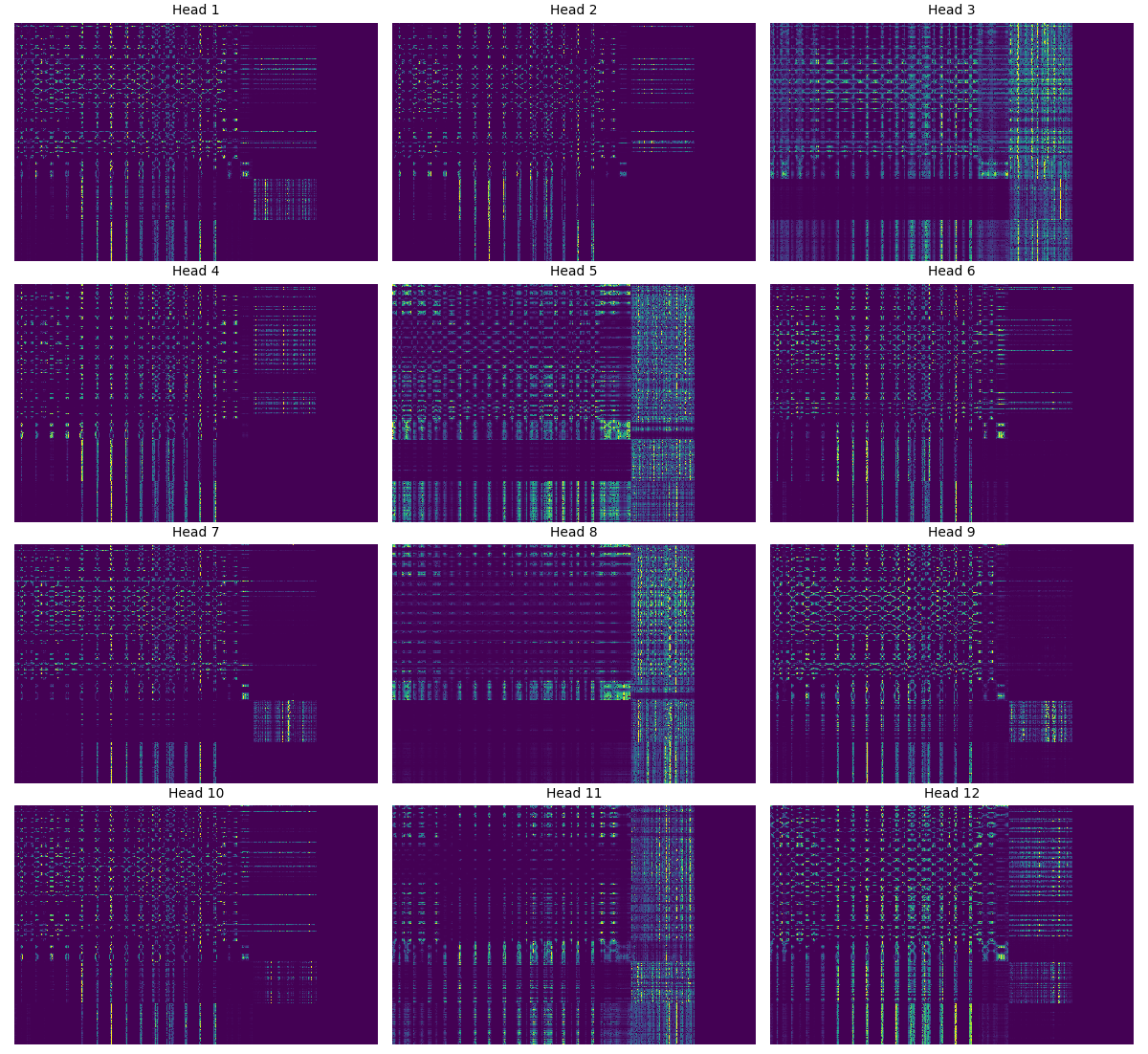}
    \caption{Visualization of Layer 12 Attention Map} 
\end{figure}

\begin{figure}
    \centering
\includegraphics[width=.85\columnwidth]{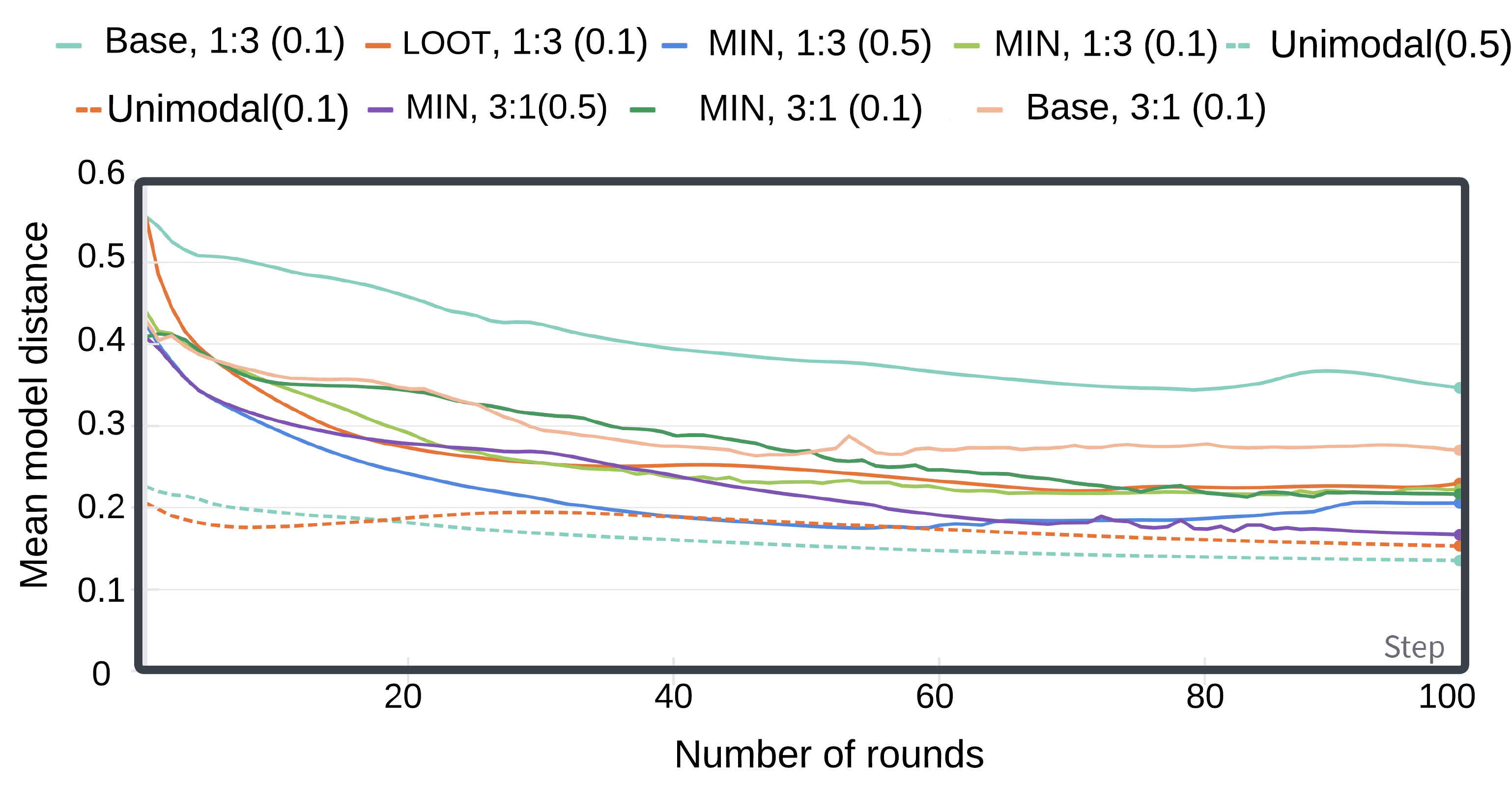}
    \caption{Mean Inter-client model L2 distance under different settings for MIMIC CXR} 
\label{fig:15}
\end{figure}

\begin{figure}
    \centering
\includegraphics[width=.85\columnwidth]{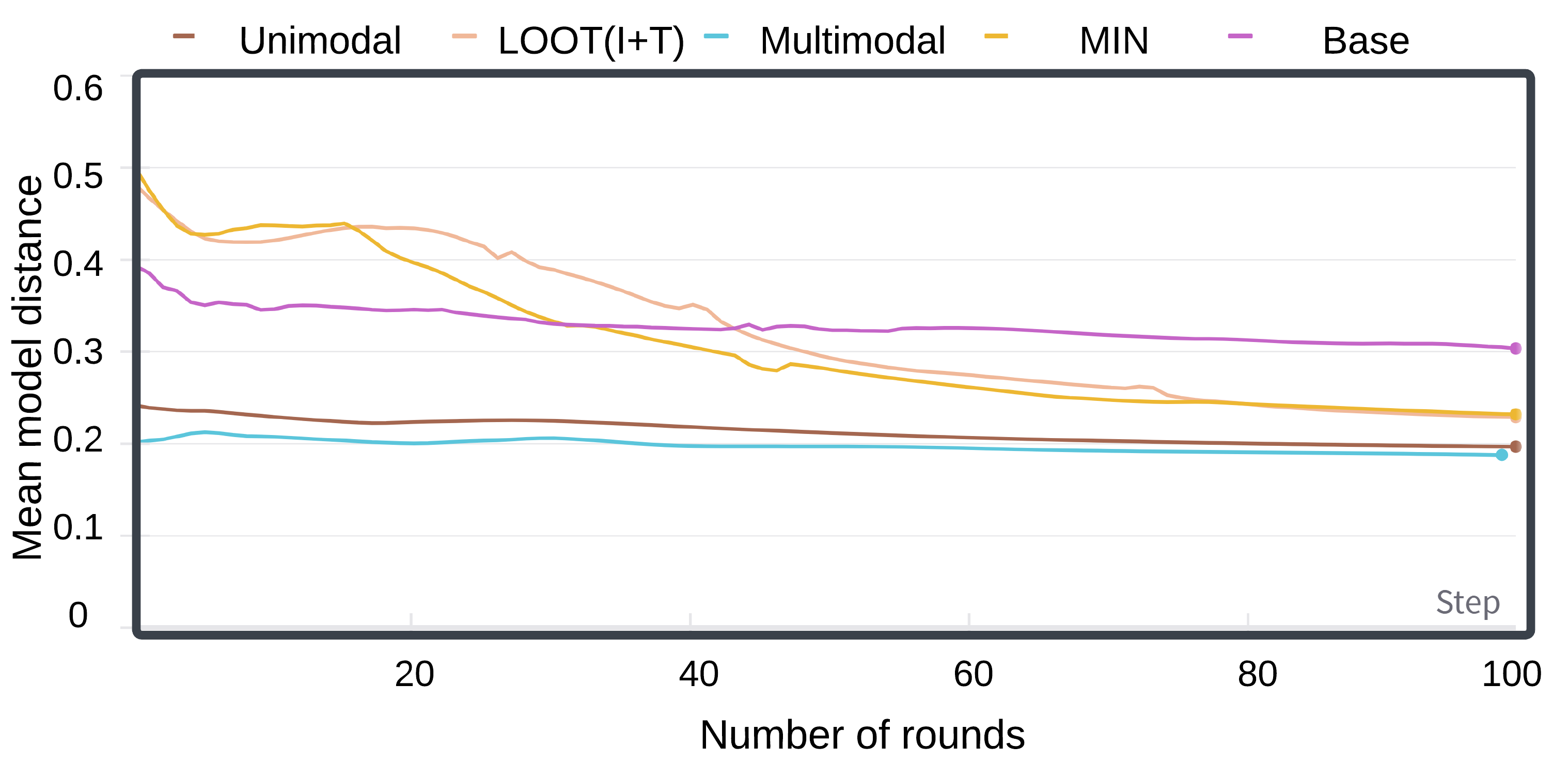}
    \caption{Mean Inter-client model L2 distance of selected models under different settings for Open-I} 
\label{fig:18}
\end{figure}

\begin{figure}
    \centering
\includegraphics[width=.95\columnwidth]{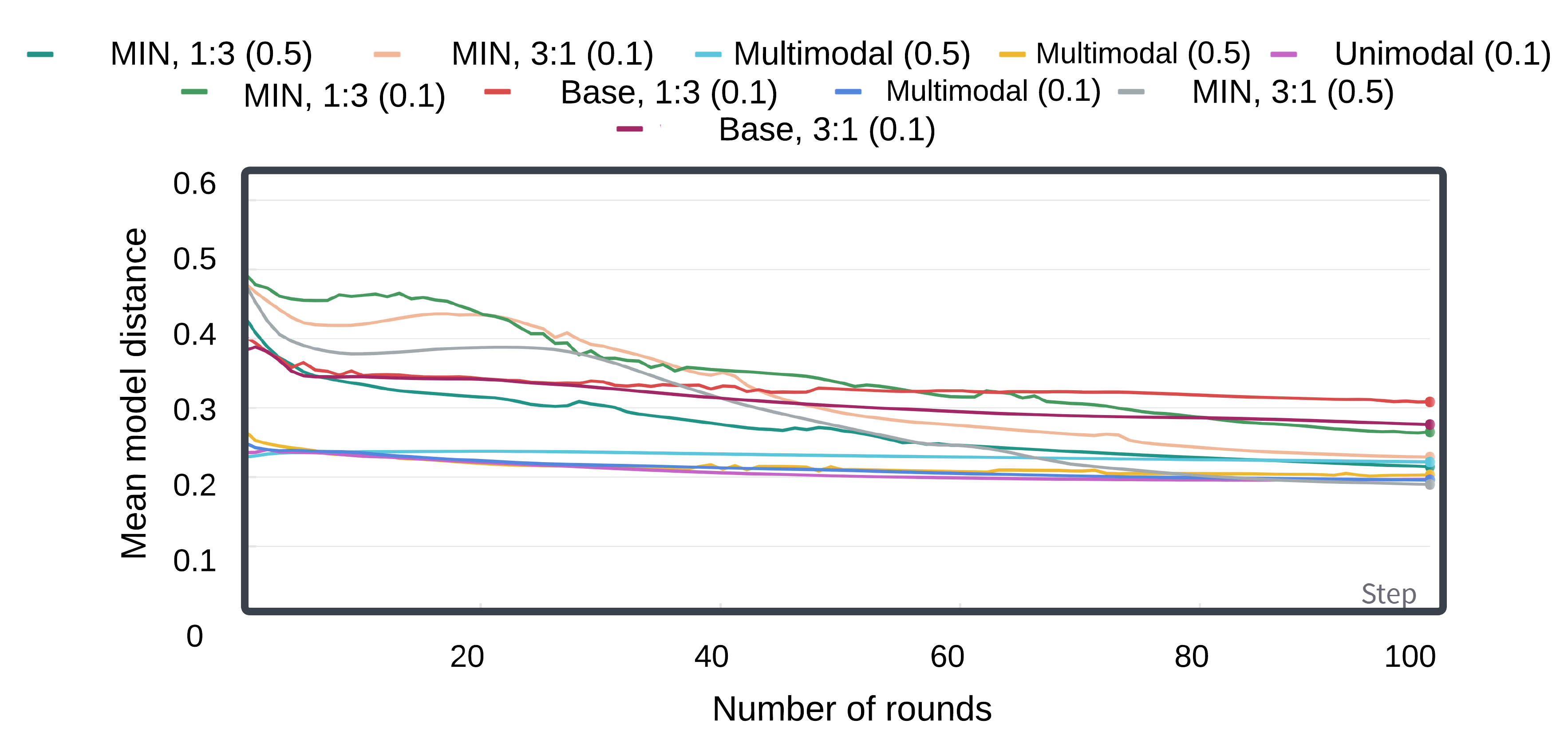}
    \caption{Mean Inter-client model L2 distance of selected models under different settings for Open-I} 
\label{fig:17}
\end{figure}

\begin{figure}
    \centering
\includegraphics[width=.75\columnwidth]{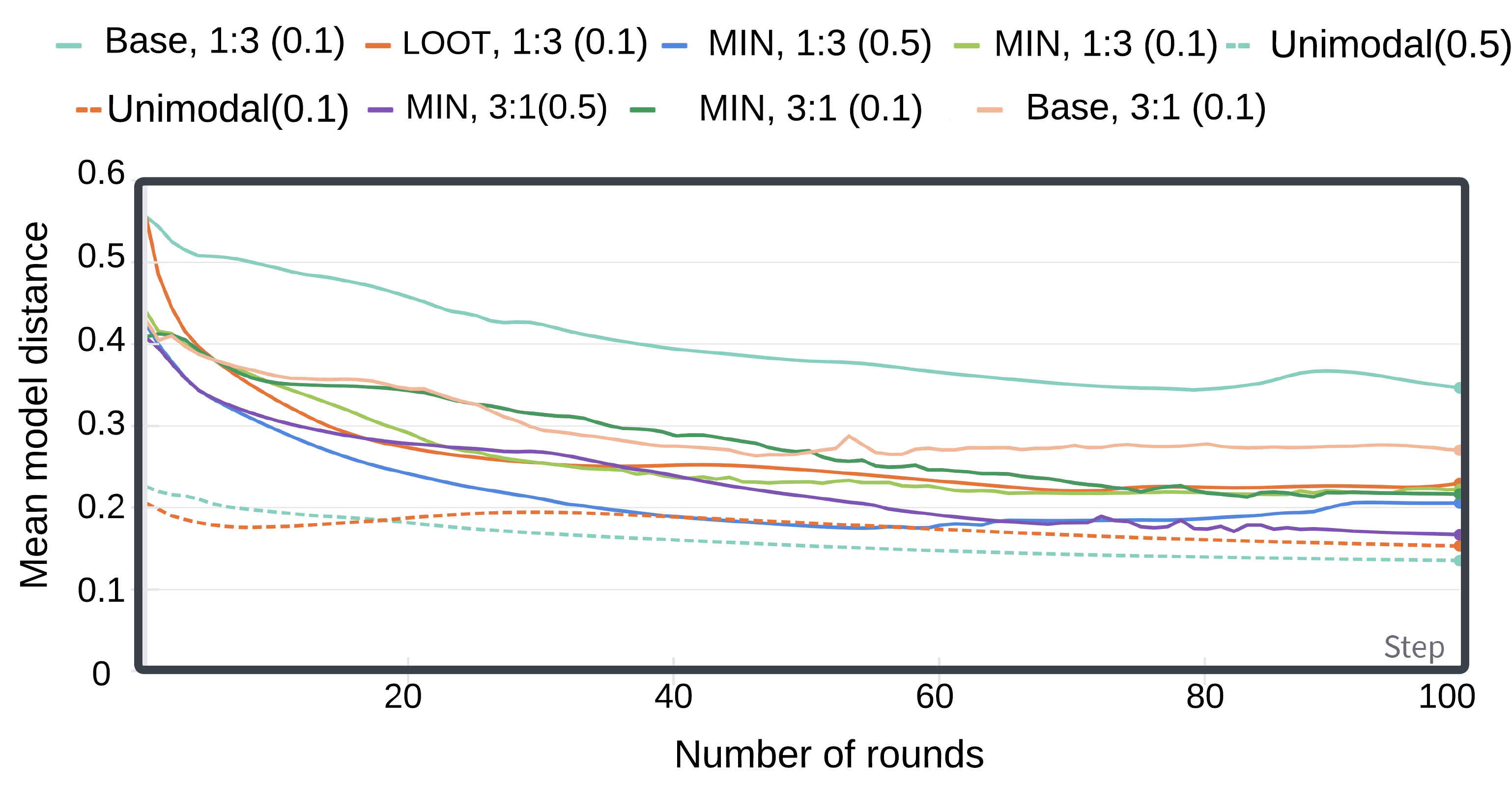}
    \caption{Per-class accuracy of MIN for MIMIC CXR. Each of the 14 curves denote the accuracy of a particular category} 
\label{fig:19}
\end{figure}

\section{Limitations, Discussions, and Future Directions}

We discuss the main results as well as the key points of the paper, along with the shortcomings and the future directions below:

\begin{itemize}
    \item \textbf{Why is our incongruent multimodal federated learning setting particularly interesting or challenging?}

Our work is particularly interesting as \color{blue}we treat the primary modality required for decision making in medical context, \textit{i.e.}, the radiology report, as the missing modality. \color{black} The Chexpert labeler-based disease labels are extracted directly from the reports which clearly shows the strong correlation between text report and disease detection. Identifying the diseases from the report is much easier than that from the CXR images as examining the images needs radiological expertise whereas the disease information is more easily accessible from the text. Hence, in general, the network tends to give more emphasis on the text than the image embeddings, if no constraints are imposed. However, when the primary modality is missing, it forces the network to extract more information from the image. This creates a challenging setting in incongruent multimodal federated learning as the network still tends to focus more on text in the multimodal client where both XCR and reports are available whereas it only attends to the image information in the unimodal clients, where reports are not available. This is potentially the primary reason behind incongruent MMFL performing worse than unimodal FL, as in the unimodal FL settings, the model focuses on same information in all the clients. Nevertheless, as shown in Figs 26-28 and discussed earlier, the proposed methods can effectively reduce the model distances, bringing it closer to fully unimodal or fully multimodal setting in each case.

\item \textbf{Inter-client class distribution gap Vs domain gap:}

This work only addresses inter-client class distribution shift without considering domain gap between clients. Hence, MIN works well in clients 2, 3, and 4. However, if there is domain shift between client 1, where the model imputation network is trained, and other clients, where the model imputation network is utilized, the performance can be expected to decrease. However, inter-client domain differences can be addressed by leveraging source-free domain adaptation-based methods in such scenarios.

\item \textbf{Hybrid methods as future work:}

In this paper, we particularly focus on the contribution of different individual methods that have the potential to address modality incongruity. A straightforward future direction is to validate the performance of hybrid methods achieved by different combinations of each individual component. Since the self-attention mechanism, server-level and client-level solutions are independent of each other, combining multiple models could lead to more powerful methods. We do not conduct experiments on this in the current work as our objective is to determine the most effective individual algorithms for dealing with modality incongruity and analyse how that changes with variation of the multimodal and unimodal client ratio or heterogeneity level.

\item \textbf{Investigating other multimodal scenarios and architectures:}

As mentioned earlier, the work is focused on chest X-Ray and radiology report. While the experimental analysis provides us with ample evidences of the modality incongruity issue and ways to tackle it, it is worth investigating other multimodal scenarios (such as T1-weighted, T2-weighted, and FLAIR MRI image-based region of interest segmentation, vision and audio-based emotion recognition, etc.) as future works to analyse if the same behaviour is observed in other such scenarios. 

Similarly, we use a BERT-based transformer model architecture for the investigation for simplicity. More powerful multimodal architectures can be employed to ensure better processing of joint information from different modalities as a plausible next step.

\item \textbf{Availability of data on server:}

In this work, we assume the presence of similar medical data (unlabeled data) on the server for investigating server-level solutions. However, it would be interesting to explore the effects of replacing the server data with natural image data (paired image and text) or other modality imaging data as well as varying the quantity of data available on the server as future works, in case same modality data is not publicly available, which is often the setting in medical imaging scenarios.

\item \textbf{Alternative training strategies for mitigating modality incongruity:}

This work investigates different self-attention schemes, modality imputation network, federated learning-based aggregation and client-level regularization schemes towards examining the extent to which they can alleviate the modality incongruity effects. A next plausible direction would be to investigate if domain adaptation-based methods used to achieve domain invariance can also be used to achieve modality invariance in order to close the gap between unimodal and multimodal clients in federated learning settings. Similarly, another popular method called modality drop-out might be implemented during training in the multimodal client(s) for constraining the model explicitly to focus more on the modality present across clients.

\item \textbf{Other aspects of modality incongruity:}

In this work, we solely focus on the missing modality aspect of modality incongruity where we assume one particular modality is present in all the clients. This setting can be extended by relaxing the assumption and accommodating the presence or absence of any modality in different clients. In such cases, modality imputation is not a reasonable solution pathway. Instead, modality invariant representations should be investigated.

\item \textbf{Other ways of information fusion:}

We limit our investigation solely to the use of self-attention in this study instead of cross-attention between the image and text modalities in order to accommodate the missing modalities in other clients (as the model gets corrupted when cross-attention is performed with a missing modality). Other attention-based methods including cross-attention as well as other interesting methods based on early fusion and late fusion might be employed to achieve more efficient and incongruity-robust information fusion.

\item \textbf{Which methods worked and when? - A take-away :}

We plot the AUC scores of 16 different methods analysed as a part of this work for mitigating modality incongruity in Fig. 30 and 31. We arrange them in ascending order of their performance for better summarization. Fig. 30 shows that for M:U=3:1 (\textit{i.e.}, three multimodal clients and one unimodal client) with Dirichlet coefficient = 0.5, 5 methods (Causal SA, MOON, FedProx, partial-Bi SA, and MultiMOON) still perform marginally worse than the fully unimodal setting. However, 11 other methods are able to mitigate the modality incongruity and perform better than unimodal setting. LOOT (I+T) is observed to be the best model that demonstrates closest performance to fully multimodal scenario, closely followed by the Modality imputation network (MIN). However, for more heterogeneous setting ($\gamma=0.1$), none of the methods successfully outperform the fully unimodal setting. Nevertheless, the otherwise best methods (MIN and LOOT (I+T)) are observed to be performing marginally close to fully unimodal case. 

Fig. 31 shows that for M:U=1:3 (\textit{i.e.}, three unimodal clients and one multimodal client), three methods can mitigate the modality incongruity effects  under $\gamma=0.5$, \textit{viz}., FedDF (I+T), MIN, LOOT (I+T). Server-level method LOOT is seen to consistently outperform MIN, the next best candidate, for moderate heterogeneity ($\gamma=0.5$). However, for higher heterogeneity $(\gamma=0.1)$, MIN outperforms LOOT. In the current setting, modality imputation is seen to be the only method that performs better than fully unimodal setting. However, as mentioned earlier, it is worth investigating as a future work whether it can maintain its performance under clients with domain shift.

\end{itemize}

\begin{figure}
    \centering
\includegraphics[width=.95\columnwidth]{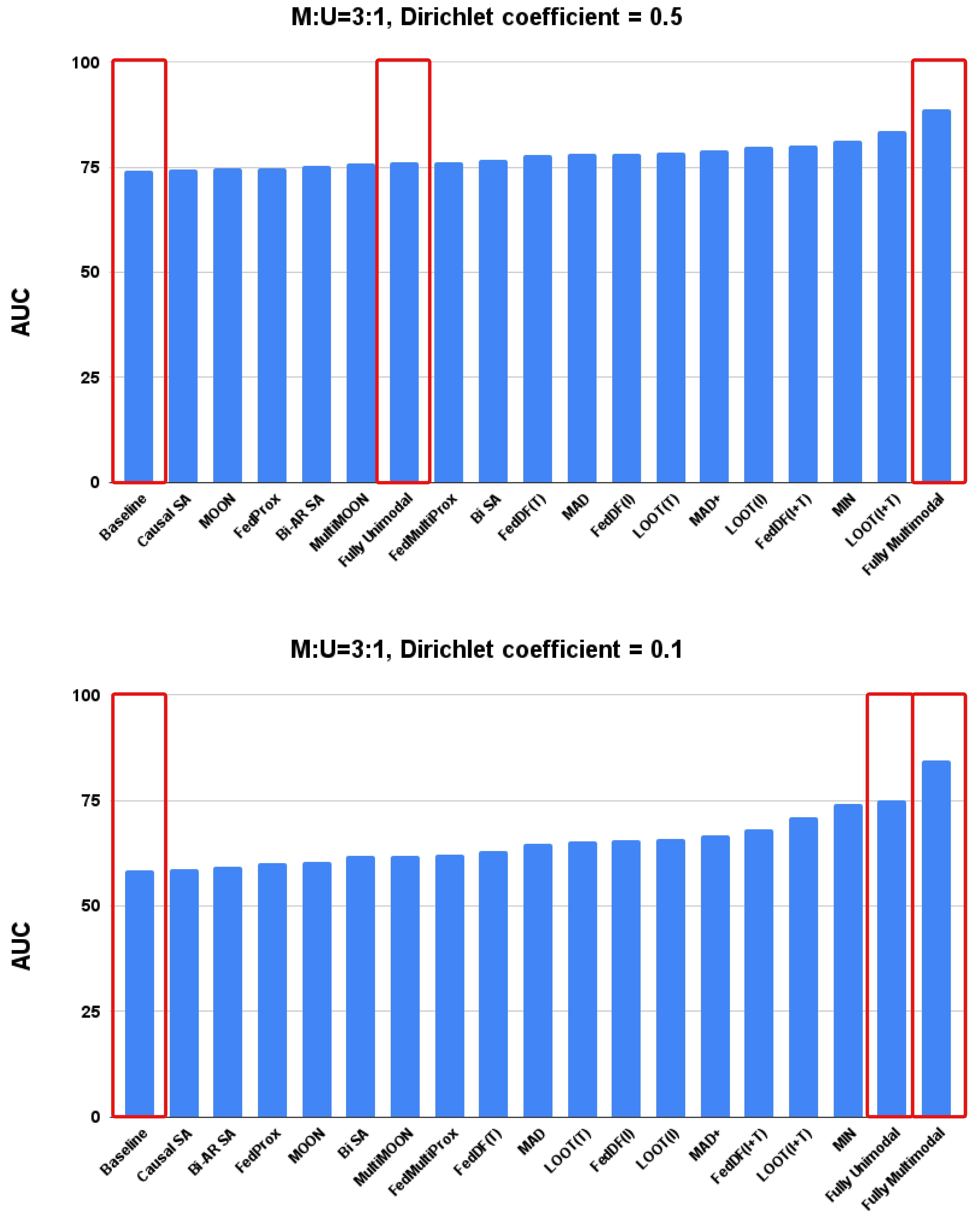}
    \caption{Comparative performance of different methods for M:U=3:1, arranged in ascending order of performance} 
\label{fig:20}
\end{figure}

\begin{figure}
    \centering
\includegraphics[width=.95\columnwidth]{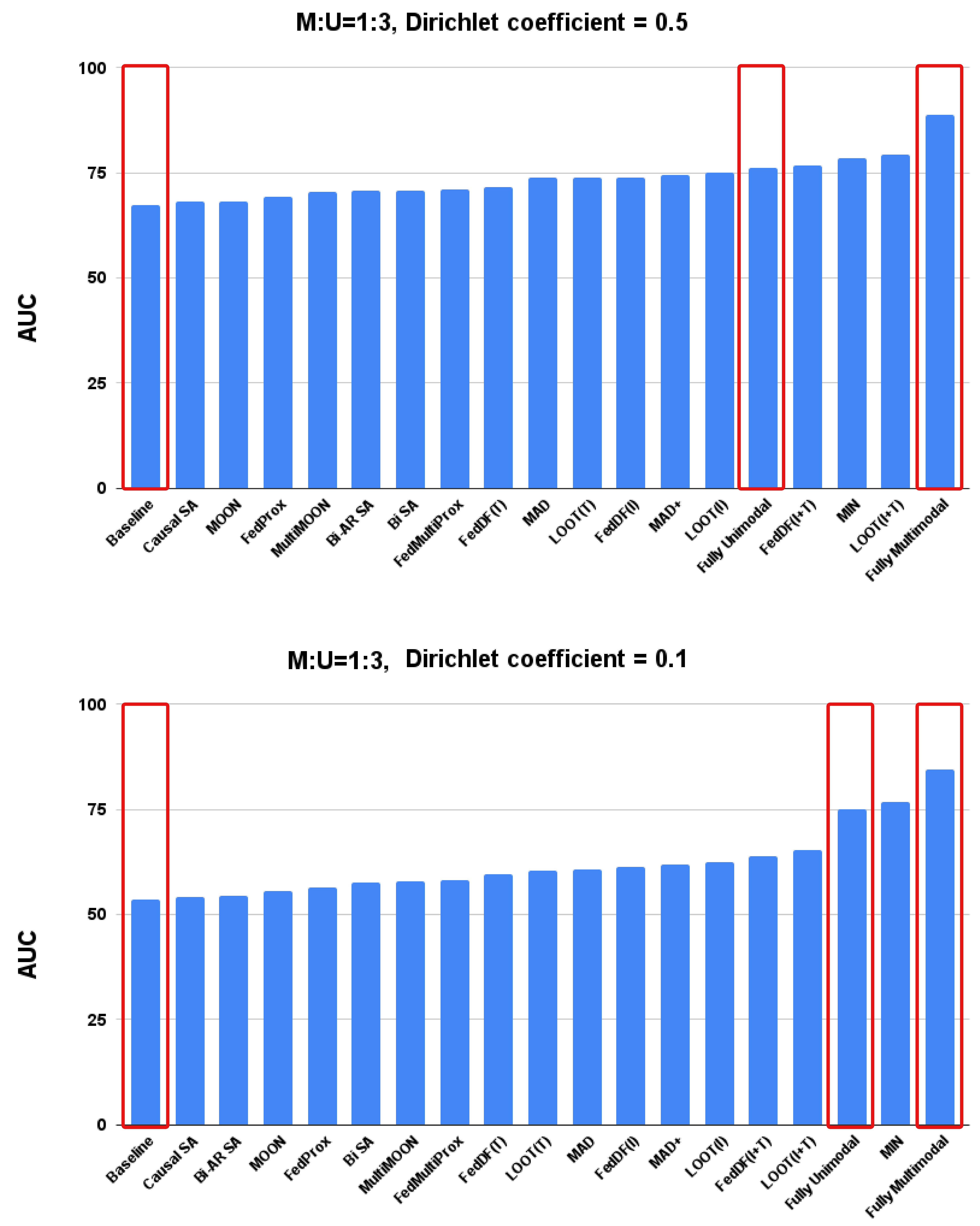}
    \caption{Comparative performance of different methods for M:U = 1:3, arranged in ascending order of performance} 
\label{fig:21}
\end{figure}


\end{document}